%% file: main.tex
\documentclass[10pt,twocolumn,letterpaper]{article}

\usepackage[pagenumbers]{cvpr} 
\usepackage{graphicx}
\usepackage{amsmath}
\usepackage{amssymb,amsthm}
\usepackage{booktabs}
\usepackage{multirow}
\usepackage{makecell}
\usepackage[dvipsnames,table]{xcolor}
\usepackage{comment}
\usepackage{xspace}
\usepackage{tabularx}
\usepackage[table]{xcolor}
\usepackage{bbding}
\usepackage{pifont}
\usepackage[accsupp]{axessibility} 

\input{preamble}

\newcommand{\OURS}{ShapeR}

\definecolor{cvprblue}{rgb}{0.21,0.49,0.74}
\usepackage[pagebackref,breaklinks,colorlinks,allcolors=cvprblue]{hyperref}

\title{\OURS: Robust Conditional 3D Shape Generation from Casual Captures}

\author{
Yawar Siddiqui~~ 
Duncan Frost~~
Samir Aroudj~~
Armen Avetisyan~~\\
Henry Howard-Jenkins~~
Daniel DeTone~~
Pierre Moulon~~
Qirui Wu{\textsuperscript{\textdagger}}~~
Zhengqin Li~\\
Julian Straub~~
Richard Newcombe~~
Jakob Engel~~
\vspace{0.2cm} \\
Meta Reality Labs Research~~~
Simon Fraser University\textsuperscript{\textdagger}~~~
\vspace{0.2cm}
}

\begin{document}
\input{figures/teaser}
\input{sections/0_abstract}    
\input{sections/1_introduction}

\input{sections/2_relatedworks}
\input{sections/3_method}
\input{sections/4_experiments}

\input{sections/5_conclusion}
{
    \small
    \bibliographystyle{ieeenat_fullname}
    \bibliography{main}
}

\input{sections/6_supplementary}


\end{document}

%% file: preamble.tex
\newcommand{\mypara}[1]{\vspace{1mm}\noindent\textbf{#1}}

%% file: figures/teaser.tex
\twocolumn[{%
	\renewcommand\twocolumn[1][]{#1}%
	\vspace{-35pt}
        \maketitle
	\begin{center}
        \vspace{-9mm}
		
        \includegraphics[width=\linewidth]{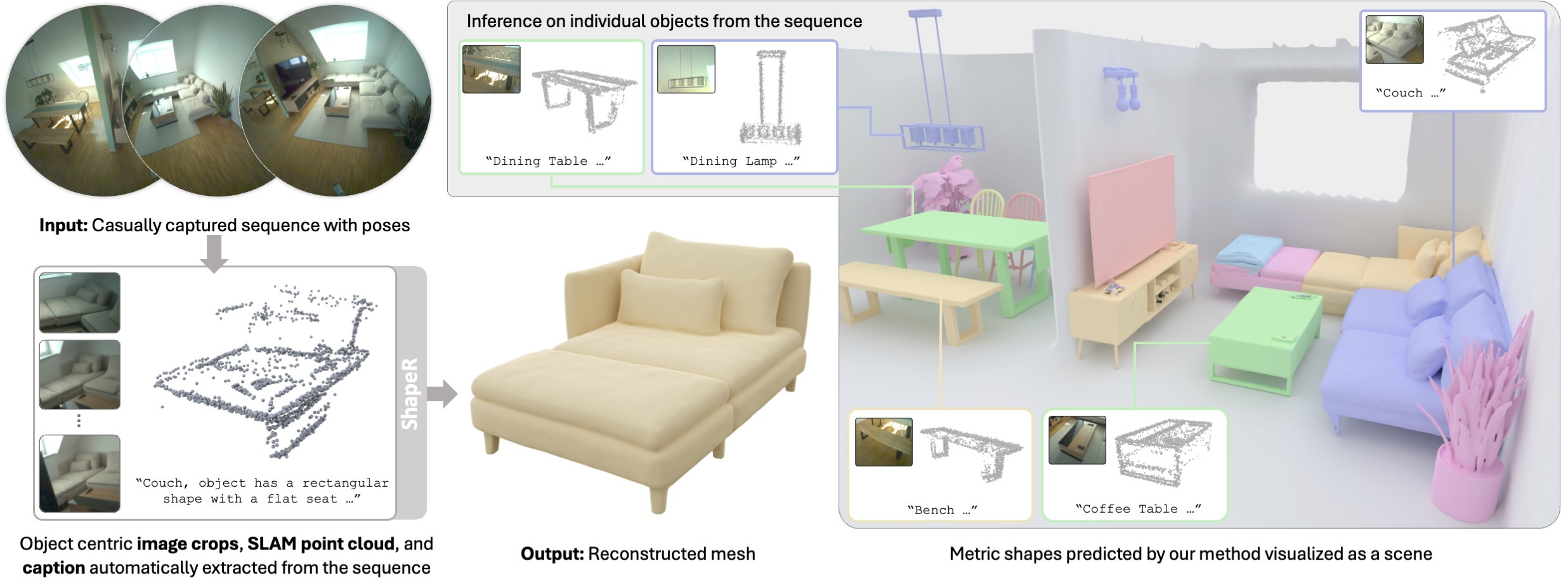}
    
		\captionof{figure}{
        ShapeR introduces a novel approach to metric shape generation. Given an input image sequence, preprocessing extracts per-object metric sparse SLAM points, images, poses, and captions using off-the-shelf methods. A rectified flow transformer operating on VecSet latents conditions on these multimodal inputs to generate a shape code, which is decoded into the object's mesh. (Right) By applying the model object-centrically to each detected object, we obtain a metric reconstruction of the entire scene.
        }
		\label{fig:teaser}
	\end{center}    
    }]

%% file: sections/0_abstract.tex
\begin{abstract}
Recent advances in 3D shape generation have achieved impressive results, but most existing methods rely on clean, unoccluded, and well-segmented inputs. Such conditions are rarely met in real-world scenarios. 
We present \OURS\footnotemark, a novel approach for conditional 3D object shape generation from casually captured sequences. 
Given a image sequence, we leverage off-the-shelf visual-inertial SLAM, 3D detection algorithms and VLMs to extract for each object, a set of sparse SLAM points, posed multi-view images, and machine-generated captions. 
A rectified flow transformer trained to effectively condition on these modalities then generates high-fidelity metric 3D shapes.
To ensure robustness to the challenges of casually captured data, we employ a range of techniques including on-the-fly compositional augmentations, a curriculum training scheme spanning object- and scene-level datasets, and strategies to handle background clutter. 
Additionally, we introduce a new evaluation benchmark comprising 178 in the wild objects across 7 real-world scenes with geometry annotations. 
Experiments show that \OURS~significantly outperforms existing approaches in this challenging setting, achieving an improvement of 2.7$\times$ in Chamfer distance compared to SoTA.
\vspace{-7mm}
\end{abstract}

%% file: sections/1_introduction.tex
\section{Introduction}
\label{sec:intro}

3D reconstruction is a longstanding challenge in computer vision, essential for understanding and interacting with the physical world.\footnotetext{\href{https://facebookresearch.github.io/ShapeR/}{facebookresearch.github.io/ShapeR}}
Scene-centric methods typically reconstruct entire scenes as single entities~\cite{mildenhall2021nerf,peng2020convolutional,straub2024efm3d,dai2021spsg,yu2022monosdf,zhu2022nice}, but produce monolithic representations, often with limited resolution and missing surfaces in unobserved areas.
Object-centric reconstruction~\cite{nie2020total3dunderstanding,huang2025midi,yao2025cast,meng2025scenegen,wu2025amodal3r,avetisyan2019scan2cad} instead focuses on recovering individual objects within a scene, enabling more detailed and complete results.

Recent advances in object-level generative models~\cite{zhao2025hunyuan3d,xiang2025structured,wu2025amodal3r,wu2025direct3d,li2025triposg}, enabled by improved architectures~\cite{dosovitskiy2020image,peebles2023scalable,esser2024scaling}, large-scale 3D datasets~\cite{deitke2023objaverse, deitke2023objaverseorg}, and better shape representations~\cite{zhang20233dshape2vecset,xiang2025structured}, have rapidly advanced object-centric shape generation. These models produce high-fidelity shapes from clean, well-segmented, and unoccluded inputs.
However, their performance drops significantly in casual capture settings, \ie, real-world scenarios with natural, non-scanning trajectories where users move freely and captures often include occlusions, background clutter, sensor noise, low resolution, and suboptimal views (\cref{fig:motivation})

To address these challenges, we propose \OURS, a large-scale rectified flow model for robust 3D shape generation from casually captured sequences. 
\OURS~is designed to leverage complementary information from multiple modalities, including sparse metric point clouds, multi-view posed images, and machine-generated captions. 
Given an input sequence, we first use off-the-shelf SLAM~\cite{engel2017direct} to obtain sparse point clouds and camera poses. Next, we apply 3D instance detection~\cite{straub2024efm3d} to extract object-centric crops from both images and point clouds, and generate text captions using vision-language models~\cite{meta2025llama}. 
These multimodal cues condition a flow-matching~\cite{lipman2022flow} transformer, which is trained to denoise latent VecSets~\cite{zhang20233dshape2vecset} that can be decoded into complete 3D shapes.

\input{figures/motivation}

To improve robustness, we apply extensive on-the-fly augmentations across all input modalities during training. 
Unlike prior work that relies on explicit 2D segmentation~\cite{li2025lirm,xiang2025structured,wu2025direct3d,li2025triposg}, \OURS~ learns to implicitly segment objects within images by utilizing the 3D instance points. 
Training is conducted in a two stage curriculum learning setup: the first stage uses large and diverse object-centric datasets with objects in isolation, where we address the limitations of these contrived settings through extensive point and image augmentations.
The second stage employs synthetic scene data~\cite{avetisyan2024scenescript}, which covers fewer categories but offers more realistic scenarios. This captures diverse object combinations that single-object datasets cannot model due to combinatorial complexity.

For evaluation, we introduce a new dataset of in-the-wild sequences with paired posed multi-view images, SLAM point clouds, and individually complete 3D shape annotations for 178 objects across 7 diverse scenes. 
In contrast to existing real-world 3D reconstruction datasets which are either captured in controlled setups~\cite{dong2025digital, kuang2023stanford} or have merged object and background geometries or incomplete shapes~\cite{yeshwanth2023scannet++,baruch2021arkitscenes}, this dataset is designed to capture real-world challenges like occlusions, clutter, and variable resolution and viewpoints to enable realistic, in-the-wild evaluation. 

We believe \OURS~ represents a key step toward unifying generative 3D shape modeling~\cite{zhao2025hunyuan3d,xiang2025structured,wu2025amodal3r,wu2025direct3d,li2025triposg} and metric 3D scene reconstruction~\cite{straub2024efm3d,peng2020convolutional,sayed2022simplerecon,yu2022monosdf,zhu2022nice}:
\OURS~ produces complete, high-fidelity object shapes at appropriate level of detail for each object, while preserving real-world metric consistency. We will release all code, model weights and the \OURS~evaluation dataset. 
In summary, our contributions are:
\begin{itemize}
    \item A rectified flow model for robust 3D metric shape generation from casually captured sequences, trained with a robust pipeline that combines sparse point clouds, posed images, on-the-fly cross-modal augmentations, and a two-stage curriculum for effective generalization.
	\item An evaluation dataset of causally captured sequences with paired images, SLAM points, and 3D shape annotations for 178 objects across seven scenes, enabling systematic evaluation under realistic conditions.
\end{itemize}

%% file: figures/motivation.tex
\begin{figure}[tp]
  \centering
   \includegraphics[width=0.90\linewidth]{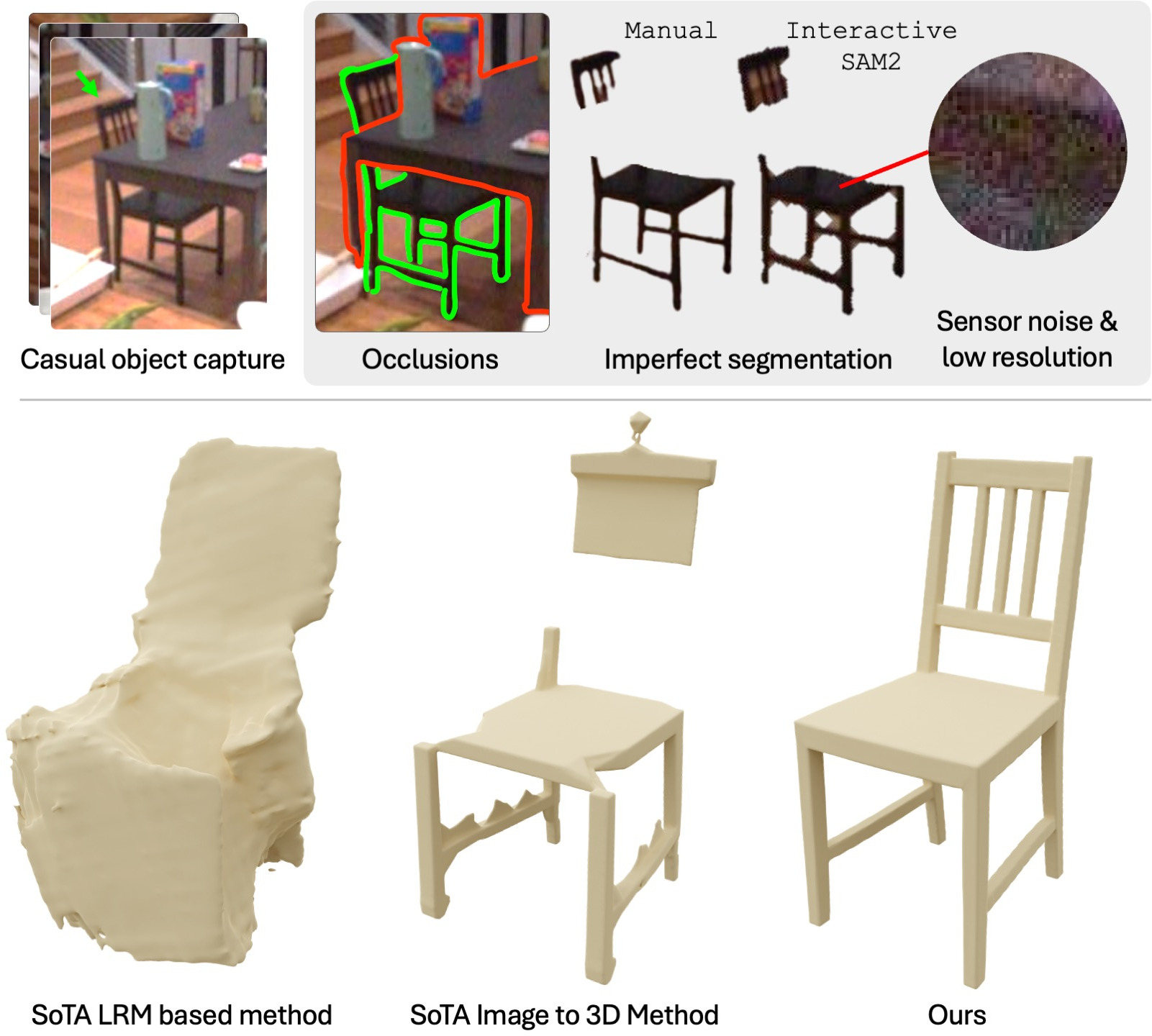}
   \vspace{-2mm}
   \caption{(Top) Objects captured in casual settings pose challenges like clutter, poor viewpoints, low resolution, noise, motion blur, and occlusions that are difficult to segment, even interactively. (Bottom) State-of-the-art 3D models often fail in these scenarios, while \OURS~remains robust and effective.}
   \label{fig:motivation}
   \vspace{-6mm}
\end{figure}

%% file: sections/2_relatedworks.tex
\section{Related Works}
\label{sec:relatedworks}

\mypara{Non Object-centric Reconstruction.} Surface reconstruction has been widely studied using both learned and optimization-based methods~\cite{kazhdan2006poisson, izadi2011kinectfusion, niessner2013real, dai2020sg, dai2021spsg, siddiqui2021retrievalfuse}. Recent approaches such as NeRF~\cite{mildenhall2021nerf}, 3DGS~\cite{kerbl20233d}, and their extensions~\cite{chen2022tensorf, muller2022instant, barron2021mip, barron2023zip, yu2024mip} achieve high-fidelity view synthesis but prioritize appearance over geometric accuracy. SDF-based implicit methods~\cite{yariv2021volume,li2023neuralangelo, wang2021neus, wang2023neus2, yariv2023bakedsdf} improve geometric faithfulness while maintaining view quality. Feedforward methods~\cite{murez2020atlas, sun2021neuralrecon, wang2025vggt, zhu2022nice, straub2024efm3d} directly predict global scene geometry from posed images, reducing optimization overhead. However, these methods reconstruct scenes as a single surface, leaving individual objects incomplete under occlusion. In contrast, \OURS~ performs explicit object-level reconstruction from sequences, producing complete geometry for each object.

\mypara{Conditional Object Reconstruction.}
Early work explored class-specific reconstruction models~\cite{park2019deepsdf, mescheder2019occupancy, peng2020convolutional, chibane2020implicit} conditioned on images or point clouds. Later methods, such as Dreamfusion~\cite{pooledreamfusion} and its extensions~\cite{wang2023prolificdreamer, chen2023fantasia3d, liu2023zero1to3, lin2023magic3d, jun2023shap}, used 2D diffusion models for text-conditioned shape generation, moving beyond fixed classes. Large Reconstruction Models~\cite{hong2023lrm} and follow-ups~\cite{siddiqui2024meta, tang2024lgm, gao2024cat3d, li2025lirm} scaled image-to-3D reconstruction and integrated mesh generation, texturing, and relightable assets, relying on 2D diffusion priors. With large-scale datasets~\cite{deitke2023objaverse}, native 3D diffusion approaches~\cite{zhang2024clay, xiang2025structured, zhao2025hunyuan3d, li2025triposg, li2025sparc3d, wu2025direct3d} have further improved fidelity. However, most methods require clean, well-segmented inputs and lack metric grounding from single images, and even amodal approaches~\cite{wu2025amodal3r} struggle in real-world scenarios. ShapeR differs by leveraging multimodal conditioning with sparse metric point clouds, posed images, and captions, enabling robust, metrically accurate reconstruction under occlusion, clutter, and viewpoint variation.

\mypara{Object-centric Scene Reconstruction.} Early approaches addressed object-centric scene reconstruction through joint detection and completion~\cite{hou2020revealnet, dahnert2021panoptic,runz2020frodo} or CAD model retrieval~\cite{avetisyan2019scan2cad, kuo2020mask2cad, avetisyan2019end}, but often resulted in incomplete or mismatched geometry. Later methods~\cite{nie2020total3dunderstanding, liu2022towards, dahnert2024coherent} reconstructed individual objects and scene layouts from single views, but were typically limited to specific classes. Recent work~\cite{ardelean2025gen3dsr, huang2025midi, yao2025cast, yang2025instascene, ni2025decompositional, meng2025scenegen} leverages diffusion priors, open-vocabulary detection, and generative models to improve per-object geometry and scene assembly, but often depends on high-quality 2D instance segmentation. While ShapeR focuses on object-centric rather than joint scene reconstruction, it generates 3D metric shapes conditioned on point cloud crops from off-the-shelf detectors, which can be composed for scene-level reconstruction. Unlike prior methods that degrade with machine-generated segments in real-world scenarios, ShapeR remains robust to imperfect segmentation and challenging, casual capture conditions.

%% file: sections/3_method.tex
\section{Method}
\label{sec:method}
ShapeR performs generative, object-centric 3D reconstruction from image sequences by leveraging multimodal inputs and robust training strategies. First, a sparse 3D point cloud and camera poses are extracted using an off-the-shelf visual-inertial SLAM method~\cite{engel2017direct}. Object instances are then identified via a 3D instance detection method~\cite{straub2024efm3d}, leveraging both SLAM points and posed images. For each detected object, its sparse points, the images in which it appears, 2D projections of its 3D points in those images, and a machine-generated caption from a vision-language model~\cite{meta2025llama} are extracted. These multimodal inputs condition a 3D rectified flow matching model, which denoises a latent VecSet~\cite{zhang20233dshape2vecset} and decodes it to produce the object's 3D shape~(\cref{fig:method_overview}). The use of multimodal conditioning, along with heavy on-the-fly compositional augmentations and curriculum training, ensures the robustness of ShapeR in real-world scenarios.
\input{figures/method_overview}

\subsection{Multimodally Conditioned Flow Matching}
Following recent advances in 3D generative modeling~\cite{xiang2025structured,zhang2024clay,zhao2025hunyuan3d,li2025triposg}, ShapeR formulates object-centric shape generation as a rectified flow process that denoises latent representations learned by a 3D VAE.

\mypara{3D Variational Autoencoder.} We adopt the Dora~\cite{chen2025dora} variant of VecSets~\cite{zhang20233dshape2vecset} as our latent autoencoder. Given a mesh $S$, two point clouds are sampled: (i) uniformly distributed surface points capturing overall geometry and (ii) edge-salient points capturing fine detail. These are separately cross-attended, downsampled, concatenated, and further processed through self-attention to produce a latent code $ z \in \mathbb{R}^{L \times d} $, where $L$ is variable in $\{256, 512, \dots, 4096\}$ and feature width $d=64$. The decoder $D$ predicts signed distance values $s=D(z, x)$ for a grid of query points $x \in \mathbb{R}^3$ through cross-attention with the processed latent sequence. The VAE is trained using
\begin{equation}
\mathcal{L}_{\text{VAE}} =  || s - s_{GT} ||_2^2 + \beta \mathcal{L}_{\text{KL}}\Big(q(z|S) \;||\;\mathcal{N}(0,I)\Big).    
\end{equation}

\mypara{Rectified Flow Model.} The latent distribution $z \sim q(z|S)$ serves as the target distribution for flow matching. A denoising transformer $f_\theta$ is trained to transport Gaussian noise $z_1 \sim \mathcal{N}(0,I)$ to the latent manifold $z_0$, conditioned on multimodal cues ( C ):
\begin{equation}
 \dot{z}_t = f_\theta(z_t, t, C), \quad t \in [0,1].   
\end{equation}
The training objective minimizes the expected squared error between the model-predicted and true transport velocity:
\begin{equation}
    \mathcal{L}_{\text{FM}} = \mathbb{E}_{t,z_t,C} \big[ || f_\theta(z_t,t,C) - (z_0 - z_1) ||_2^2 \big].
\end{equation}
We employ a FLUX.1-like dual-single-stream transformer~\cite{batifol2025flux}, where the first four dual layers cross-attend to text tokens and subsequent dual and single layers to image and point tokens. Similar to~\cite{zhao2025hunyuan3d,li2025triposg}, positional embeddings are omitted. Dual-stream outputs are concatenated and subsequently processed by several self-attention layers. Both dual and single stream blocks are modulated with timestep and CLIP~\cite{radford2021learning} text embeddings.

\input{figures/point_ablation}

\input{figures/augmentation_example}

\mypara{Condition Encoding.} Condition inputs $ C = \{ C_{\text{pts}}, C_{\text{img}}, C_{\text{txt}} \} $ comprise of the 3D SLAM points, images, and captions respectively. For $C_{\text{pts}}$, a ResNet~\cite{he2016deep} style 3D sparse-convolutional encoder downscales the point features into a token stream. For $C_{\text{img}}$, a frozen DINOv2~\cite{oquab2023dinov2} backbone extracts image tokens, concatenated with Pl\"ucker ray encodings of the corresponding camera poses. The object's 3D points observed in their respective frames are projected to 2D to form binary point masks, which are processed by a 2D convolutional extractor and concatenated with DINO and Pl\"ucker tokens. For $C_{\text{txt}}$, captions are tokenized with a frozen T5 encoder~\cite{raffel2020exploring} and a CLIP~\cite{radford2021learning} text encoder. Notably, no segmentation masks are used; the object of interest is learned implicitly from the 3D point tokens and 2D projected point mask information.

\subsection{Two-Stage Curriculum Learning Setup} As a class-agnostic generative model, ShapeR must learn priors across diverse categories. In the first stage, we train on a large-scale object-centric dataset containing over 600K meshes of diverse semantic categories created by 3D artists. To simulate noisy, real-world inputs, we apply extensive augmentations to all modalities (\cref{fig:augmentation_example}, left), including background compositing, occlusion overlays, visibility fog, resolution degradation, and photometric perturbations on images. For SLAM points, we simulate partial trajectories, a diverse range of point dropout strategies, Gaussian noise, and point occlusion. These augmentations are applied on-the-fly in the data loader in a compositional manner, yielding a virtually infinite stream of unique training samples.
While this stage teaches the model general shape priors, it does not fully reflect the complexity of real captures. Hence, we fine-tune the model on a second dataset consisting of object crops extracted from Aria Synthetic Environments~\cite{avetisyan2024scenescript}. Although this dataset is less diverse, it exhibits realistic occlusions, inter-object interactions, and SLAM noise patterns (\cref{fig:augmentation_example}, right).

\subsection{Inference} Given a posed image sequence $I = {I^1,\dots, I^K}$ and corresponding camera intrinsics \& extrinsics $\Pi = {\Pi^1, \dots, \Pi^K}$, we first compute sparse metric point clouds $P$ by tracking and triangulating high-gradient image regions similarly to \cite{engel2014lsd}. This provides both 3D point positions and their visibility association across frames, represented as $P_{I^k} \subseteq P$, denoting the subset of points observed in frame $I^k$.
An instance detection model~\cite{straub2024efm3d} is applied on the posed images and point cloud to predict 3D bounding boxes for object instances. For each object $i$, the corresponding point set $P_i \subset P$ is refined within its bounding box using SAM2~\cite{ravi2024sam} to remove spurious samples from neighboring instances. Using the point–frame association $P_{I^k}$, we identify all frames where object $i$ is visible and select a fixed number $N$ of representative frames $I_i$. For each selected frame $I_{i}^{j}$, the points $P^i \cap P_{I_{i}^{j}}$ are projected onto the image plane to generate binary masks $M_i$, approximating the object’s silhouette in that view. A vision–language model~\cite{meta2025llama} is then prompted on each object's representative image to generate a descriptive caption $T_i$. The complete conditioning set for object $i$ is thus $C_i = \{ P_i, I_i, \Pi_i, M_i, T_i \}$.
Before generation, each object’s point cloud $P_i$ is normalized to the normalized device coordinate cube $[-1,1]^3$. The flow-matching model predicts the object’s shape within this normalized space, and the reconstructed mesh is rescaled back to the original metric coordinate system of $P_i$, ensuring physically accurate dimensions. Sampling proceeds by integrating the learned flow:
\begin{equation}
 z_1 \sim \mathcal{N}(0,I), \quad z_{t-\Delta t} = z_t + \Delta t\,f_\theta(z_t, t, C_i),   
\end{equation} with midpoint sampling. The final mesh is reconstructed as 
\begin{equation}
\hat{S}_i = \text{Rescale}\big(\text{MarchingCubes}(D(z_0)), P_i\big),    
\end{equation}
producing metrically consistent, fully reconstructed meshes for each detected object $i$, aligned with the real-world scale and placement of the input sequence.

\mypara{Implementation Details.} The point cloud is derived from images using SLAM or SfM; specifically, we use semi-dense point clouds from Project Aria’s Machine Perception Services~\cite{engel2023project}, obtained via a visual-inertial SLAM system with Aria’s monochrome cameras and IMUs. During training, conditioning is performed using two randomly sampled views per object, while inference uses up to sixteen selected views at a resolution of $280\times280$ pixels from Aria Mono scene SLAM cameras. Additional details are provided in the Appendix \cref{sec:spl_implementation}.

%% file: figures/method_overview.tex
\begin{figure}[htp]
  \centering
   \includegraphics[width=\linewidth]{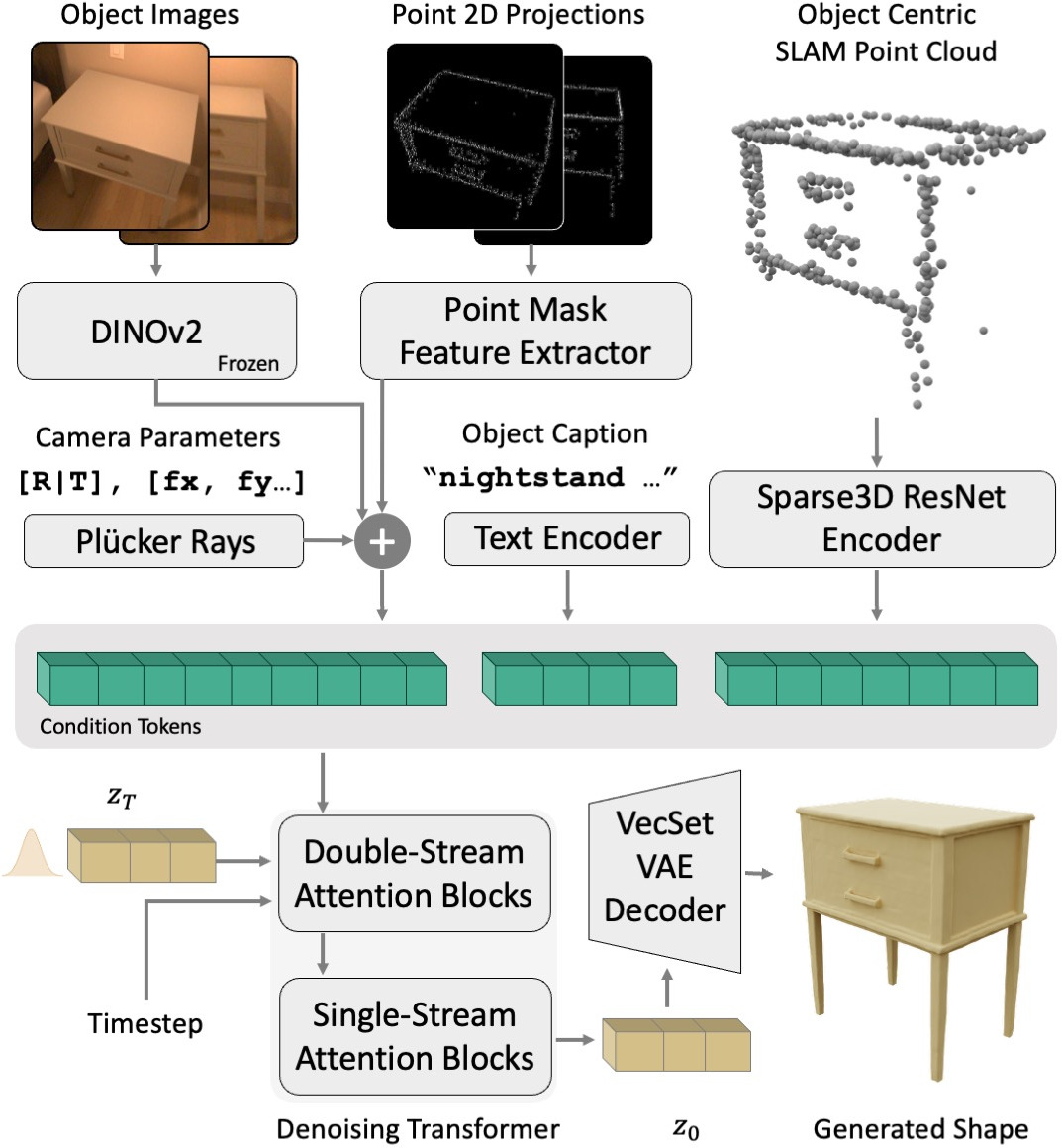}
   \caption{The ShapeR denoising transformer, built on the FLUX DiT architecture, denoises latent VecSets by conditioning on multiple modalities: posed images, SLAM points, captions, and the 2D projections of SLAM points observed in those input images. SLAM points are encoded with a sparse 3D ResNet, images using a frozen DINOv2 backbone, poses using Pl\"ucker encodings, and projection masks via a 2D convolutional network. The denoised latent is decoded into a SDF, from which the final object shape is extracted using marching cubes.}
   \label{fig:method_overview}
   \vspace{-4mm}
\end{figure}

%% file: figures/point_ablation.tex
\begin{figure}[tp]
  \centering
   \includegraphics[width=0.98\linewidth]{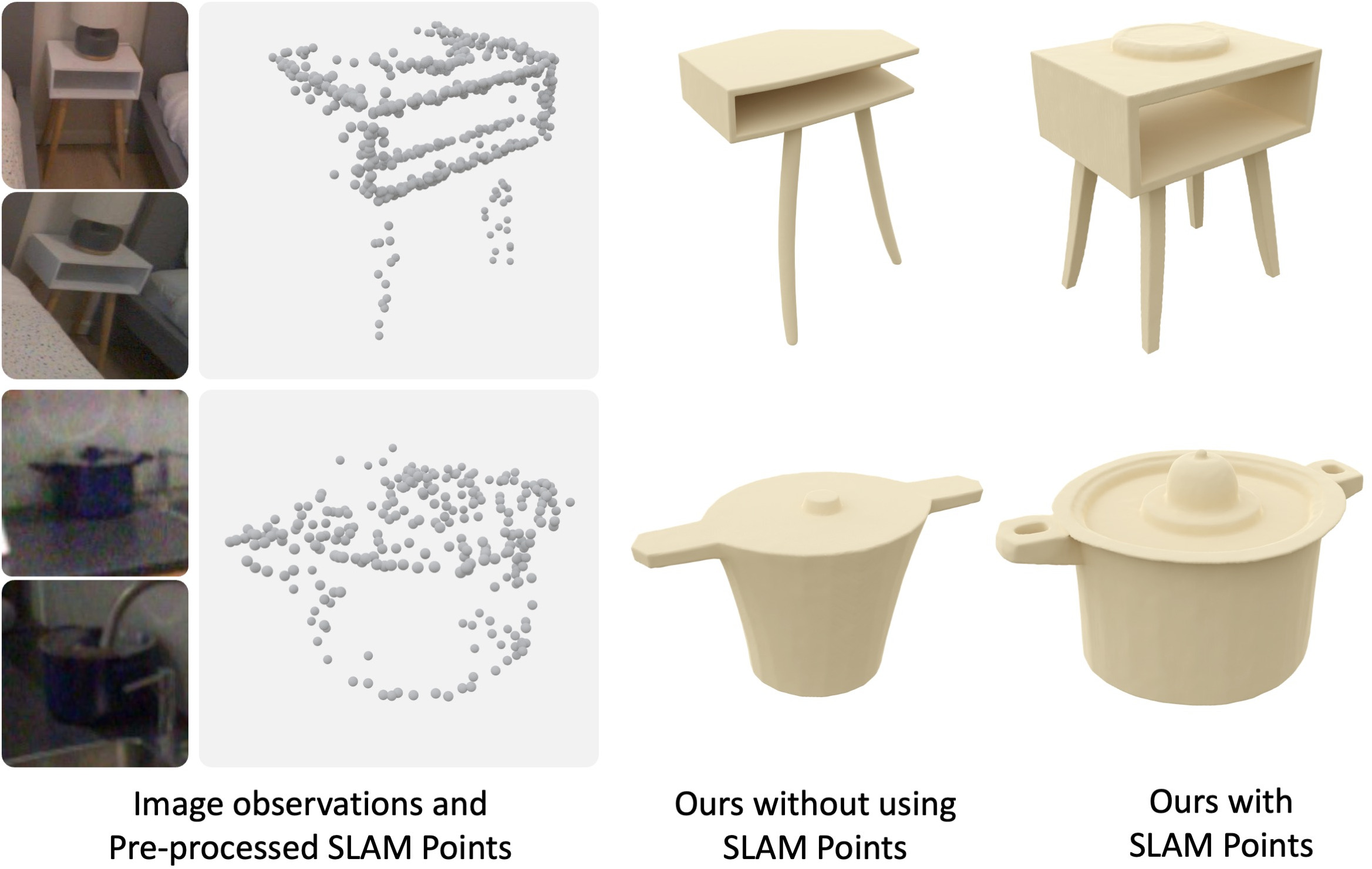}
   \vspace{-2mm}
   \caption{Incorporating SLAM points significantly enhances robustness. These points provide a complementary geometric signal to posed images, encoding aggregated shape information across the entire sequence.}
   \label{fig:point_ablation}
   \vspace{-4mm}
\end{figure}

%% file: figures/augmentation_example.tex
\begin{figure*}[htp]
  \centering
   \includegraphics[width=\linewidth]{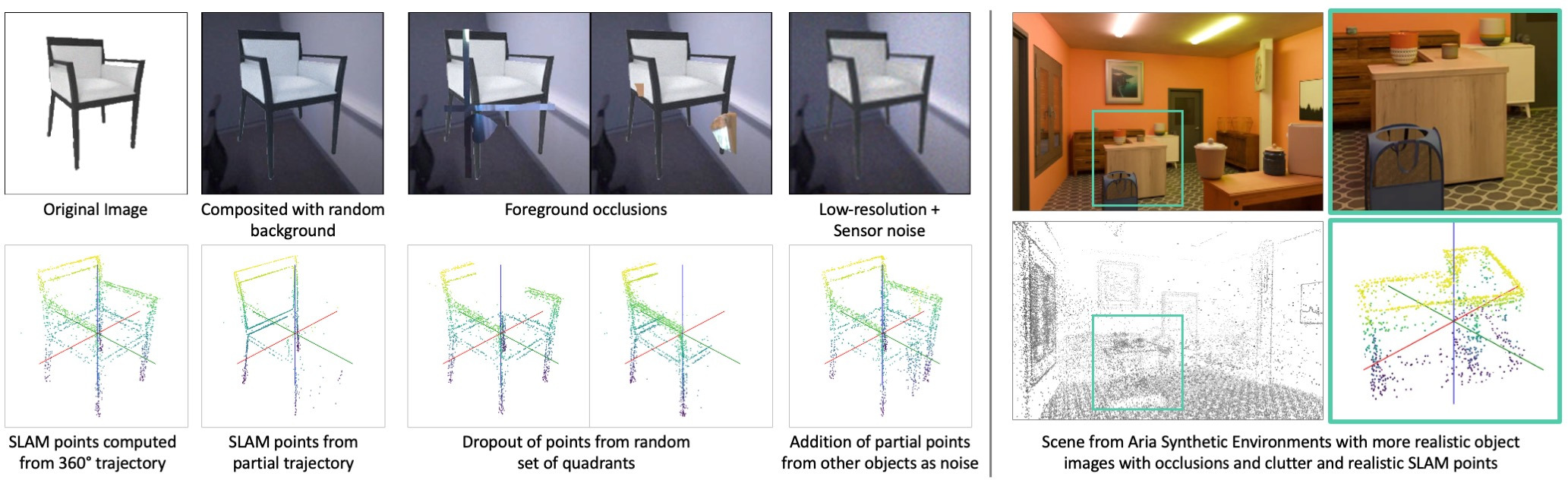}
   \vspace{-6mm}
   \caption{(Left) We pretrain on 600K object meshes with extensive, compositional augmentations across all modalities, simulating realistic backgrounds via image compositing, and introducing diverse occlusions and noise in both images and SLAM points. (Right) We then fine-tune on object-centric crops from Aria Synthetic Environment scenes, which feature realistic image occlusions, SLAM point cloud noise, and inter-object interactions.}
   \label{fig:augmentation_example}
   \vspace{-5mm}
\end{figure*}

%% file: sections/4_experiments.tex
\section{Experiments}
\label{sec:experiments}

We evaluate and ablate key components of ShapeR on a novel high quality dataset against nine leading 3D reconstruction and generation methods, grouped by the type of input they exploit and the nature of their reconstruction task.

\mypara{ShapeR Evaluation Dataset.}
While several datasets exist for benchmarking 3D reconstruction~\cite{dong2025digital, downs2022google, dai2017scannet, chang2015shapenet, deitke2023objaverse, yeshwanth2023scannet++, baruch2021arkitscenes, kuang2023stanford}, most are limited in realism or completeness. Synthetic datasets such as ShapeNet~\cite{chang2015shapenet} and Objaverse~\cite{deitke2023objaverse,deitke2023objaverseorg} offer large-scale coverage but lack real-world complexity. Controlled datasets like DTC~\cite{dong2025digital}, GSO~\cite{downs2022google}, and StanfordORB~\cite{kuang2023stanford} focus on isolated tabletop objects in studio settings. In-the-wild datasets such as ScanNet~\cite{dai2017scannet}, ScanNet++~\cite{yeshwanth2023scannet++}, and ARKitScenes~\cite{baruch2021arkitscenes} provide realistic imagery but lack complete object-level 3D geometry for evaluation (\cref{fig:dataset_motivation}).
To address these gaps, we introduce the ShapeR Evaluation Dataset, designed to benchmark reconstruction under challenging, casual capture conditions. 
The dataset contains seven casually-captured recordings from distinct cluttered scenes annotated with 178 diverse high quality object shapes.
It covers a wide range of categories, from large objects like furniture to smaller items such as remotes, toasters, and tools as can be seen in \cref{fig:results_3drecon,fig:results_imageto3d,fig:dataset_annotation,fig:dataset_example}. For each sequence, we provide multi-view images, calibrated camera parameters, SLAM point clouds, and machine-generated object captions. Each annotated object also includes a complete reference mesh generated using internal image-to-3D modeling methods under ideal conditions, which we manually refined and realigned for geometric and pose consistency. More details are provided in the supplementary. All quantitative and qualitative evaluations in the following sections are conducted on this dataset. Evaluations on further datasets are in the Appendix~\cref{sec:spl_experiments}.

\mypara{Metrics.} We evaluate the reconstructed geometry with $3$ complementary metrics  following prior works~\cite{peng2020convolutional,siddiqui2024meta,siddiqui2021retrievalfuse}: Chamfer $\ell_2$ Distance (CD), Normal Consistency (NC) and F-score (F1) at $1\%$ threshold. All metrics are computed in the normalized coordinate space. 

\subsection{Results}
\input{figures/results_3drecon}
\input{tables/reconstruction_comparison}

\mypara{Posed Multi-view to 3D.} We compare against EFM3D~\cite{straub2024efm3d}, TSDF fusion with FoundationStereo depths~\cite{wen2025foundationstereo}, DP-Recon~\cite{ni2025decompositional}, and LIRM~\cite{li2025lirm}. These methods take posed images and predict metric 3D geometry. For monolithic mesh predictors such as EFM3D and FoundationStereo-based fusion, we extract object instances by cropping the predicted mesh using ground-truth geometry as guidance. For DP-Recon and LIRM, which rely on 2D object segmentations, we provide SAM2-generated masks. As shown in \cref{tab:quantitative_results} and \cref{fig:results_3drecon}, monolithic scene reconstruction methods produce incomplete objects due to occlusions, while segmentation-based methods degrade under imperfect masks. ShapeR, by contrast, reconstructs complete, metric shapes without requiring segmentation inputs, remaining robust across casual captures.

\mypara{Foundation Image to 3D.} We also evaluate against recent large-scale image-to-3D generative models including TripoSG~\cite{li2025triposg}, Direct3DS2~\cite{wu2025direct3d}, Hunyuan3D-2.0~\cite{zhao2025hunyuan3d}, and Amodal3R~\cite{wu2025amodal3r}. Hunyuan3D-2.0, TripoSG and Direct3DS2 are trained to predict shapes from one or multiple unposed views and perform well under idealized, clean conditions with minimal occlusion. Amodal3R, which extends TRELLIS~\cite{xiang2025structured}, improves robustness by reasoning about occluded regions and generating amodal completions. We found that for non-standard viewpoints common in casual captures, their single-view versions are significantly more competitive than the multi-view ones, so we report results using the single-view setting. To ensure their optimal performance, we manually select views with clear object visibility and use interactive SAM2-based segmentations, while ShapeR operates fully automatically using multiple posed views. Our method achieves metrically consistent, complete, and robust reconstructions without any manual intervention as shown in \cref{tab:user_study} and \cref{fig:results_imageto3d}.
\input{tables/userstudy}
\input{figures/results_imageto3d}

\mypara{Image to Scene Layout.}
We also compare against scene-level reconstruction methods, MIDI3D~\cite{huang2025midi} and SceneGen~\cite{meng2025scenegen}, which predict multiple object geometries and spatial layout. MIDI3D uses a single image, while SceneGen takes multiple views; both require interactive instance segmentation. Although effective in simplified settings, these methods struggle with realistic, cluttered scenes, often yielding inconsistent object scales and layouts (\cref{fig:results_image2scene}). In contrast, ShapeR reconstructs objects automatically with consistent scale and layout. Comparison to the recent SAM3D Objects is provided in Appendix \cref{sec:spl_experiments}.
\input{figures/results_image2scene}
\input{figures/results_ablation}

\subsection{Ablation Study of \OURS~Components}

\mypara{Effect of SLAM Points.}
We evaluate the impact of adding SLAM points as an input modality. As shown in \cref{tab:quantitative_results} and \cref{fig:point_ablation}, while image-only inputs yield reasonable reconstructions, incorporating SLAM points significantly improves robustness by providing complementary geometric information that encodes aggregated shape across the entire sequence, especially benefiting cases with weak visual cues.

\mypara{Effect of Augmentations.}
\cref{tab:quantitative_results} and \cref{fig:results_aug_ablation}(a) show that both point cloud and image augmentations are critical for robust real-world performance. Removing either leads to degraded reconstructions under noise and partial observations. The variant without image augmentation relies on explicit foreground segmentation, similar to foundation image-to-3D models, and therefore struggles with noisy masks, underscoring the importance of synthetic occlusion and background augmentation over mask dependence.

\mypara{Effect of Two-stage Curriculum Training.}
Fine-tuning on a more realistic scene dataset substantially improves robustness as shown in \cref{tab:quantitative_results} and \cref{fig:results_aug_ablation}(b). This confirms that combining large-scale object-centric pretraining with realistic scene fine-tuning provides strong generalization to casual captures.

\mypara{Effect of 2D Point Mask Prompting.}
Without the 2D point mask cues, our method sometimes reconstructs adjacent objects. Using 2D point masks to guide DINO features mitigates this issue and leads to cleaner reconstructions, as illustrated in \cref{fig:results_aug_ablation}(c) and \cref{tab:quantitative_results}.

%% file: figures/results_3drecon.tex
\begin{figure*}[htp]
  \centering
   \includegraphics[width=0.95\linewidth]{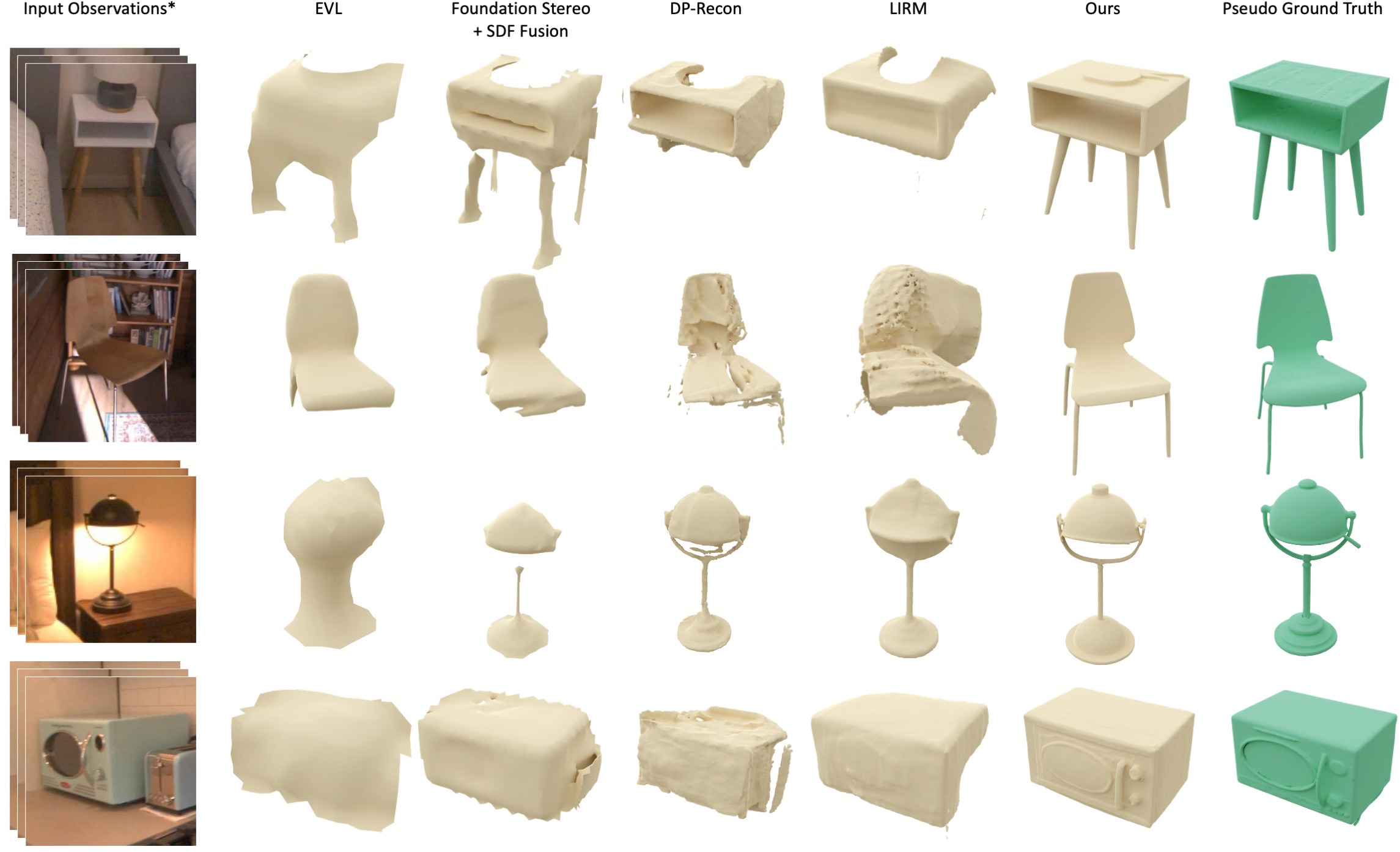}
   \vspace{-2mm}
   \caption{Qualitative comparison on the ShapeR evaluation dataset against posed multiview-to-3D methods. For scene-centric fusion approaches (EVL, Foundation Stereo), ground-truth meshes are used to segment individual object shapes. For methods relying on image segmentation masks (DP-Recon, LIRM), we employ SAM2, prompted with bounding boxes, to generate input image masks.}
   \label{fig:results_3drecon}
   \vspace{-4mm}
\end{figure*}

%% file: tables/reconstruction_comparison.tex
{
	\begin{table}[tb]
		\begin{center}
			\small
			\resizebox{0.95\linewidth}{!}{
				\begin{tabular}{lcrr}
					\toprule
					Method                   & CD$\downarrow$ \tiny{$\times 10^{2}$} & NC$\uparrow$  & F1$\uparrow$   \\
					\midrule
					EFM3D~\cite{straub2024efm3d}          & 13.82          & 0.614          & 0.276 \\
					FStereo~\cite{wen2025foundationstereo}   & 6.483          & 0.677          & 0.435       \\
					LIRM~\cite{li2025lirm}   & 8.047          & 0.683          & 0.384          \\
					DP-Recon~\cite{ni2025decompositional}   & 8.364          & 0.661          & 0.436       \\
					\midrule
					w/o SLAM Points          & 4.514          & 0.765          & 0.486          \\
					w/o Point Augmentation   & 3.276          & 0.805          & 0.667          \\
					w/o Image Augmentation   & 3.397          & 0.778          & 0.649          \\
					w/o Two Stage Training   & 3.053          & 0.801          & 0.689          \\
					w/o Point Mask Prompting & 2.568          & \textbf{0.813}          & 0.701          \\
					\cmidrule{1-4}
					\OURS{}                  & \textbf{2.375} & 0.810 & \textbf{0.722} \\
					\bottomrule
				\end{tabular}}
			\caption{Comparison on ShapeR evaluation dataset against posed multiview to 3D approaches, and an ablation of components.}
			\label{tab:quantitative_results}
		\end{center}
        \vspace{-8mm}
	\end{table}
}

%% file: tables/userstudy.tex
{
	\begin{table}[tb]
		\begin{center}
			\small
			\resizebox{0.575\linewidth}{!}{
				\begin{tabular}{lr}
					\toprule
					Method                   & ShapeR Win Rate$\uparrow$   \\
					\midrule
					  TripoSG        & 86.67\%\\
					Amodal3R    & 86.11\%\\
					Direct3DS2   & 88.33\%\\
                    Hunyuan3D-2.0   & 81.11\%\\
					\bottomrule
				\end{tabular}}
                \vspace{-2mm}
			\caption{Percentage of users who prefer our method over the image-to-3d baselines over 660 responses. Our generated meshes are preferred significantly more often.}
			\label{tab:user_study}
		\end{center}
        \vspace{-9mm}
	\end{table}
}

%% file: figures/results_imageto3d.tex
\begin{figure*}[htp]
  \centering
   \includegraphics[width=0.95\linewidth]{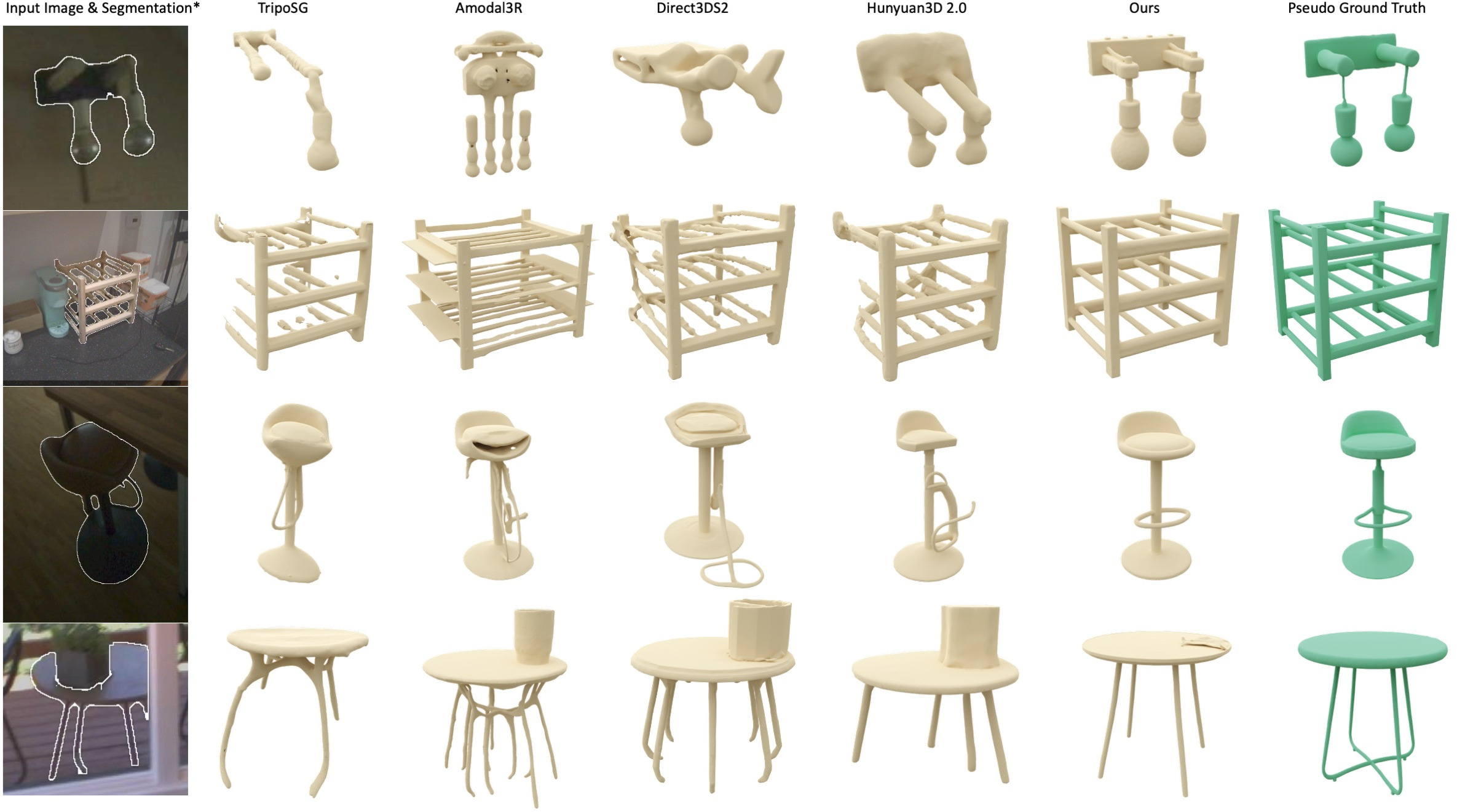}
   \caption{Qualitative comparison against foundation image-to-3D models. For these baselines, we manually select a view with clear object visibility and use interactive SAM2-based segmentations to provide optimal input. In contrast, ShapeR operates fully automatically on multiple posed views and preprocessed inputs, requiring no manual intervention.}
   \label{fig:results_imageto3d}
\end{figure*}

%% file: figures/results_image2scene.tex
\begin{figure*}[htp]
  \centering
   \includegraphics[width=\linewidth]{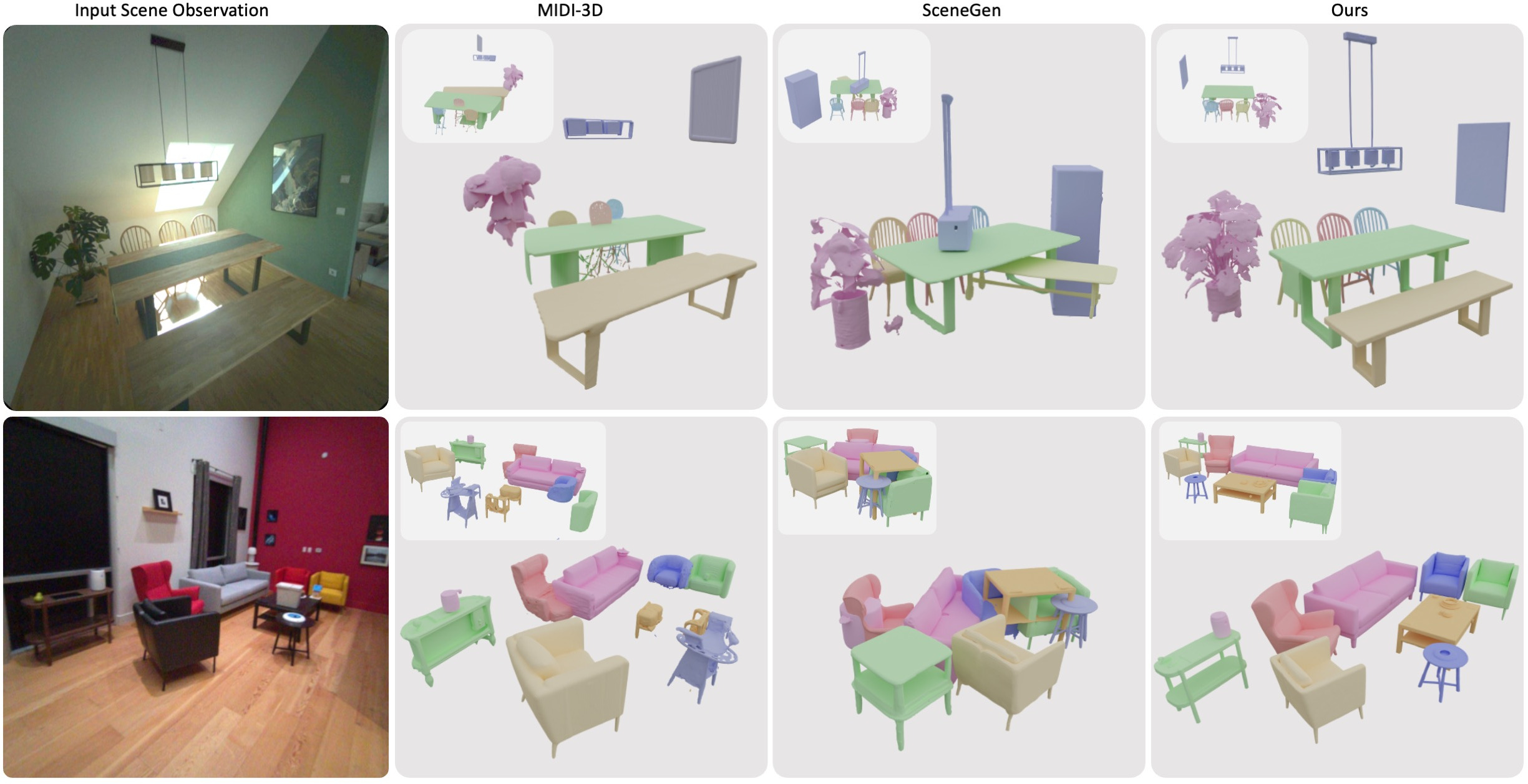}
   \caption{
   Comparison with image-to-scene methods. MIDI uses a single image and SceneGen uses four views, both with manual object segmentations. These approaches struggle with object scale and arrangement, while ShapeR reconstructs each object metrically and independently, maintaining consistent scale and layout across the scene and without interactive segmentation.
   }
   \label{fig:results_image2scene}
\end{figure*}

%% file: figures/results_ablation.tex
\begin{figure*}[htp]
  \centering
   \includegraphics[width=0.87\linewidth]{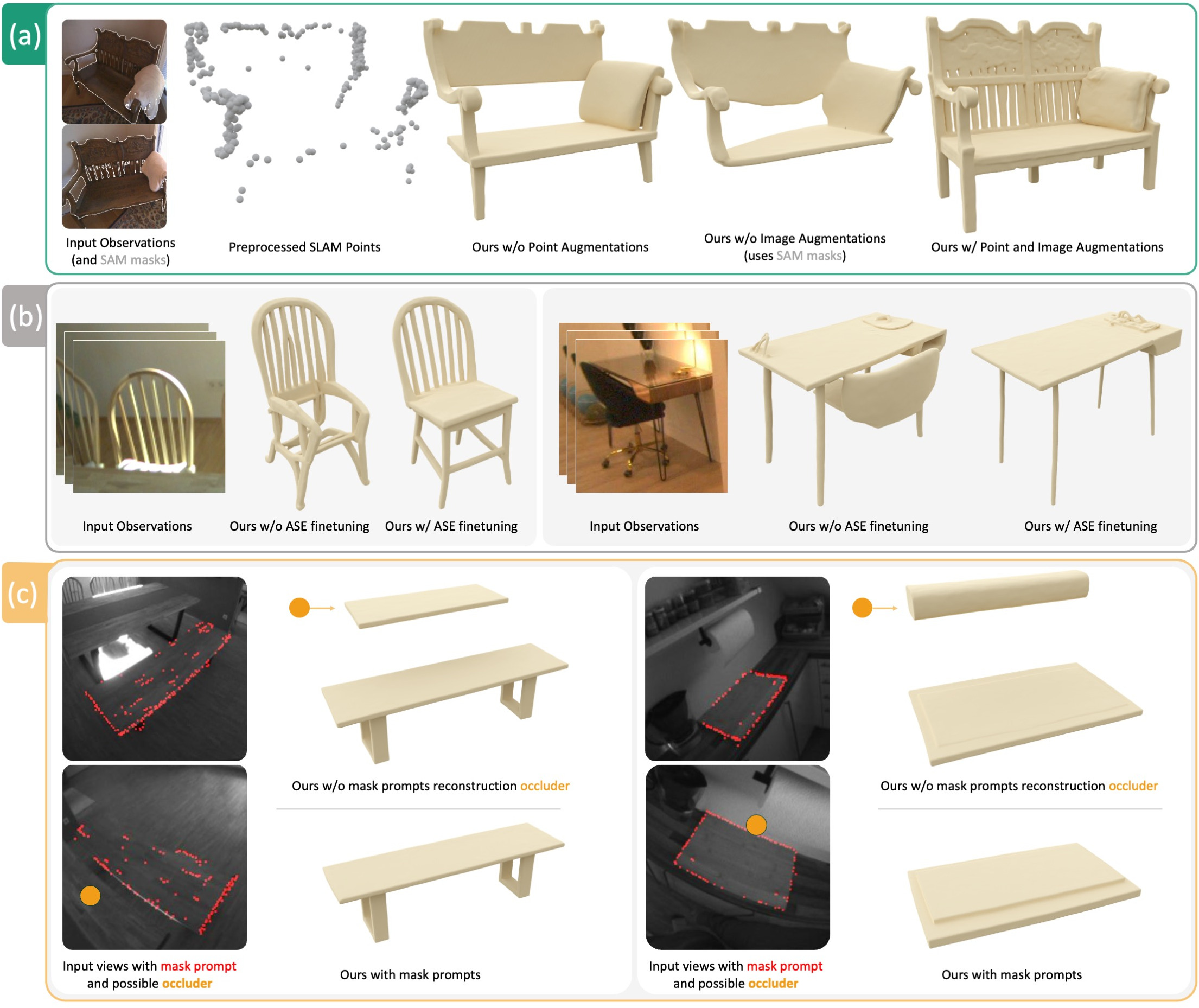}
   \vspace{-1mm}
   \caption{
   Ablations of components. (a) Without point augmentations, the model overfits to point inputs, missing geometry in regions without points. Image augmentations address occlusions and incomplete objects crops. Omitting background composition requires pre-segmentation, which can introduce noisy masks and prediction errors. (b) Fine-tuning on scene-centric crops improves robustness in challenging scenarios over object-centric training alone. (c) Prompting DINO features with 2D point projections clarifies which object to reconstruct in cluttered scenes, reducing confusion from nearby objects and improving reconstruction accuracy. 
   }
   \label{fig:results_aug_ablation}
   \vspace{-4mm}
\end{figure*}

%% file: sections/5_conclusion.tex
\section{Conclusion}
\label{sec:conclusion}
We introduce ShapeR, a multimodally conditioned rectified flow model for robust 3D shape generation from casually captured sequences. By leveraging posed images, sparse SLAM points, and textual cues, ShapeR reconstructs objects accurately and completely without explicit segmentation. Large-scale training, extensive augmentations, and a two-stage curriculum enable strong generalization to real-world scenarios. We also present the ShapeR Evaluation Dataset as a benchmark for object-centric reconstruction under casual capture. ShapeR advances scalable and automatic 3D reconstruction in natural environments.

%% file: sections/6_supplementary.tex
\maketitlesupplementary
\appendix

\noindent In this appendix, we provide additional details on the ShapeR evaluation dataset, further experimental results, including results on additional datasets, expanded implementation details of our method, and a discussion of its limitations.

\input{figures/dataset_motivation}

\input{figures/dataset_example}
\input{figures/dataset_annotation}

\input{figures/comparison_lirm_graph}

\input{figures/mapanything}

\input{figures/limitations}

\input{figures/comparison_sam3d}

\input{figures/comparison_lirm_qualitative}

\input{figures/result_thirdparty}

\section{ShapeR Evaluation Dataset}
Existing real-world 3D datasets for object reconstruction can be classified into two broad categories. Some, like Digital Twin Catalog~\cite{dong2025digital}, StanfordORB~\cite{nie2020total3dunderstanding} and Google Scanned Objects~\cite{downs2022google} provide complete 3D shape geometry, but only in highly controlled setups. Here, objects are the central focus, placed on uncluttered, disoccluded tabletops, and captured in studio-like conditions (see \cref{fig:dataset_motivation} left). These datasets typically feature relatively small objects. Others, like ScanNet~\cite{dai2017scannet}, ScanNet++~\cite{yeshwanth2023scannet++}, Matterport3D~\cite{chang2017matterport3d} offer realistic scene arrangements, with clutter and occlusions captured casually. However, these are not suitable for object-centric evaluation, as the target geometry, usually obtained by 3D scanning, is incomplete in occluded for unobservable regions (see \cref{fig:dataset_motivation} right). The ShapeR Evaluation Dataset addresses these limitations by providing complete mesh geometry annotations for a selected set of objects, while maintaining casual capture conditions. 

As shown in \cref{fig:dataset_annotation}, sequences are recorded using Project Aria \cite{engel2023project} Gen 1 or Gen 2 glasses, with the annotator casually walking through the scene and collecting images from the device’s RGB and CV cameras. Aria Machine Perception Services~\cite{engel2023project} are then used to extract SLAM points and camera parameters from the sequence. For a selected set of objects, we obtain 3D shape annotations by moving each object to an area free of clutter and occlusions, capturing a high-resolution image, and manually segmenting it. A state-of-the-art image-to-3D model is then used to generate the 3D geometry. This geometry is manually verified for plausibility and aligned to the object's position in the original casual sequence using a web interface. This interface allows annotators to reposition and rigidly deform the shape in 3D space, guided by SLAM points from the sequence. Annotators further verify placement and dimensions by projecting the mesh into the original sequence images.

In total, we annotate 178 objects across 7 real indoor sequences, spanning a range of categories. \cref{fig:dataset_example} shows sample objects and the distribution of categories in the dataset.

\section{Additional Experiments}
\label{sec:spl_experiments}
In this section, we provide additional evaluations of ShapeR across a variety of datasets and tasks. We include comparisons against SegmentAnything 3D Objects~\cite{sam3dobjects}, assessments on ScanNet++~\cite{yeshwanth2023scannet++} and Replica~\cite{straub2019replica}, results on the Digital Twin Catalog~\cite{dong2025digital} (DTC), analysis of robustness trends, and demonstrations of monocular image-to-3D reconstruction.

\mypara{Comparison against SegmentAnything 3D Object~\cite{sam3dobjects}.} SAM 3D Objects was very recently released and addresses the single image-to-3D reconstruction task using \textit{interactive} segmentation. This approach marks a significant improvement in shape quality compared to previous image-to-scene methods like MIDI3D \cite{huang2025midi} and SceneGen \cite{meng2025scenegen}, as well as single image-to-3D models such as Hunyuan3D \cite{zhao2025hunyuan3d}, Amodal3R \cite{wu2025amodal3r}, and Direct3DS2 \cite{wu2025direct3d}.

However, SAM 3D Objects is fundamentally limited by its reliance on single images. As a result, the reconstructed shapes are not metrically accurate. When scenes become more cluttered and contain multiple objects, the method struggles: layout, shape quality, aspect ratios, and relative scales all deteriorate, as shown in \cref{fig:comparison_sam3d}. In contrast, ShapeR leverages multiple posed views and additional modalities (such as SLAM points) to \textit{automatically} reconstruct objects with metric accuracy and robust layout, even in casual, cluttered environments, while having only ever been trained on synthetic data. This multimodal approach enables ShapeR to maintain high-quality, metrically consistent reconstructions and object arrangements without interaction, outperforming single image-based methods in challenging real-world scenarios. 

\mypara{Evaluation on Scannet++~\cite{yeshwanth2023scannet++} and Replica~\cite{straub2019replica}.} \cref{fig:results_thirdparty} and \cref{tab:comparison_thirdparty} present a comparison of ShapeR on third-party casually captured datasets. For these experiments, we follow the protocol of DP-Recon \cite{ni2025decompositional}, using their six ScanNet++ scenes and seven Replica scenes for evaluation. Since these datasets do not provide complete 3D geometry for evaluation (\cref{fig:dataset_motivation,fig:results_thirdparty}), we report only recall-based metrics. Notably, ShapeR produces complete reconstructions, often surpassing the ground-truth scans in terms of completeness, as the ground-truth meshes lack geometry in occluded regions.

\mypara{Evaluation on Digital Twin Catalog (DTC)~\cite{dong2025digital}.} \cref{fig:comparison_dtc_lirm} and \cref{tab:comparison_dtc} show a comparison of ShapeR against LIRM \cite{li2025lirm} on the controlled capture datasets DTC Active and DTC Passive. Both datasets contain approximately 100 sequences each, with objects placed on a tabletop, free from occlusions and clutter. The passive variant allows for more free user movement, making it more casual compared to the active variant, where the user circles the object.
As highlighted in \cref{tab:comparison_dtc}, ShapeR matches state-of-the-art LIRM quality on the highly controlled active set and surpasses it on the more casual passive variant. Additionally, ShapeR produces sharper details on both datasets, as illustrated in \cref{fig:comparison_dtc_lirm}.

\mypara{Robustness Trends.} DTC Active, DTC Passive, and the ShapeR evaluation dataset represent a non-linear progression from highly controlled to markedly more complex and casual capture setups. As shown in \cref{fig:comparison_trend}, ShapeR demonstrates significantly greater robustness to increased scene casualness compared to baseline methods such as LIRM, maintaining high reconstruction quality even as the capture conditions become more challenging.

\mypara{Monocular Image-to-3D.} While ShapeR is trained using multiple posed views and SLAM points extracted from them, it can also be applied to monocular images to produce metric 3D shapes without retraining by leveraging approaches like MapAnything \cite{keetha2025mapanything}. As illustrated in \cref{fig:map_anything}, ShapeR can condition on a single image and its associated point cloud (obtained from MapAnything) to reconstruct both individual objects and entire scenes. Further improvements are possible by fine-tuning the model on real data collected in this monocular setup, as demonstrated in recent works~\cite{sam3dobjects}.

\input{tables/thirdparty}
\input{tables/comparison_dtc}

\section{Implementation Details}
\label{sec:spl_implementation}
The 3D VAE encoder consists of $8$ transformer layers and the decoder of $16$ layers, each with a hidden width of $768$, $12$ attention heads. The VAE is trained for $200$K steps with an effective batch size of $640$ across $64$ NVIDIA H$100$ GPUs. The rectified flow transformer comprises $16$ dual-stream and $32$ single-stream blocks, each with $16$ attention heads and a hidden width of $1024$. Training is performed for $550$K steps using $128$ H$100$ GPUs, progressively increasing the latent sequence length. The effective batch size is $512$. Both networks are optimized using Adam with a learning rate of $5\times10^{-5}$. 

\section{Limitations}
While ShapeR advances 3D shape generation under casual capture scenarios, several limitations remain. First, for objects captured with low image fidelity or observed in very few views, reconstructions can be incomplete or lack fine detail due to insufficient geometric and visual evidence. Second, when objects have other items stacked or closely attached (for example, tables supporting other objects), the reconstructed meshes sometimes include remnants of these adjacent structures instead of cleanly isolating the target object. Finally, ShapeR depends on upstream 3D instance detection; thus, missed detections or inaccurate bounding boxes directly propagate to the reconstruction stage, where missed objects cannot be recovered.

%% file: figures/dataset_motivation.tex
\begin{figure}[b]
  \centering
   \includegraphics[width=\linewidth]{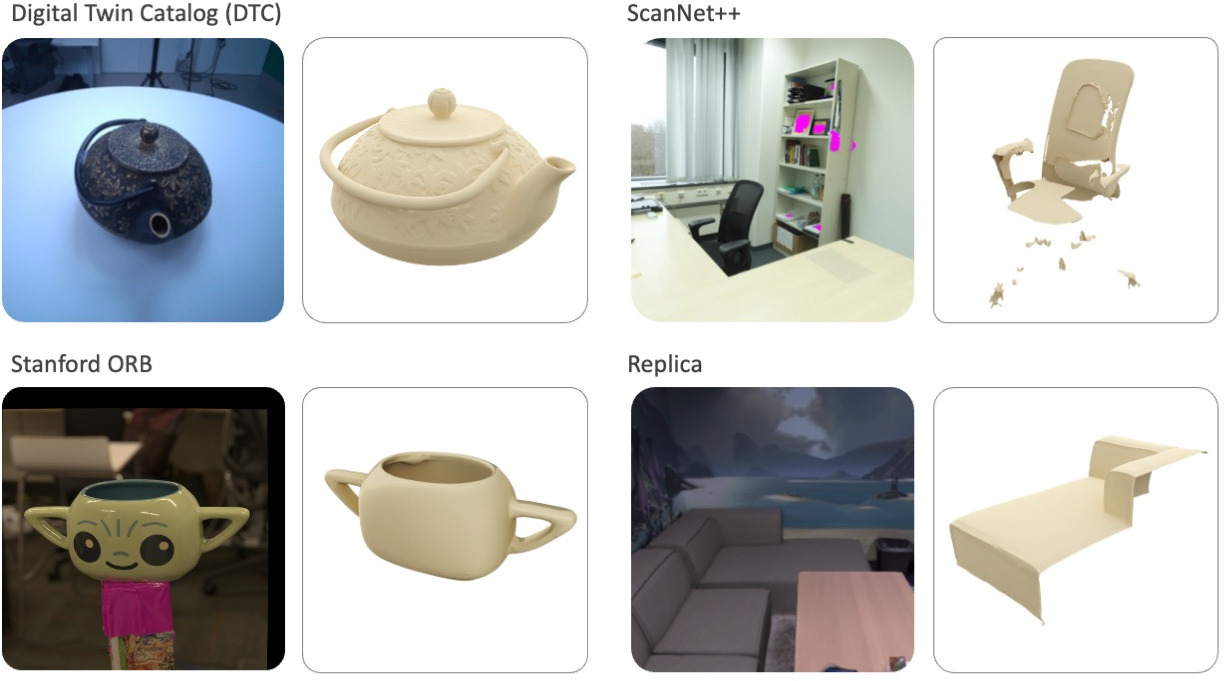}
   \vspace{-4mm}
   \caption{Comparison of 3D reconstruction datasets. DTC \cite{dong2025digital} and StanfordORB \cite{kuang2023stanford} offer controlled studio captures of isolated objects, while ScanNet++ \cite{yeshwanth2023scannet++} and Replica \cite{straub2019replica} provide realistic scenes but lack complete ground-truth shapes. The ShapeR evaluation dataset features casually captured sequences with complete meshes for geometric evaluation (see \cref{fig:dataset_example,fig:dataset_annotation}).}
   \label{fig:dataset_motivation}
\end{figure}

%% file: figures/dataset_example.tex
\begin{figure}[b]
  \centering
   \includegraphics[width=\linewidth]{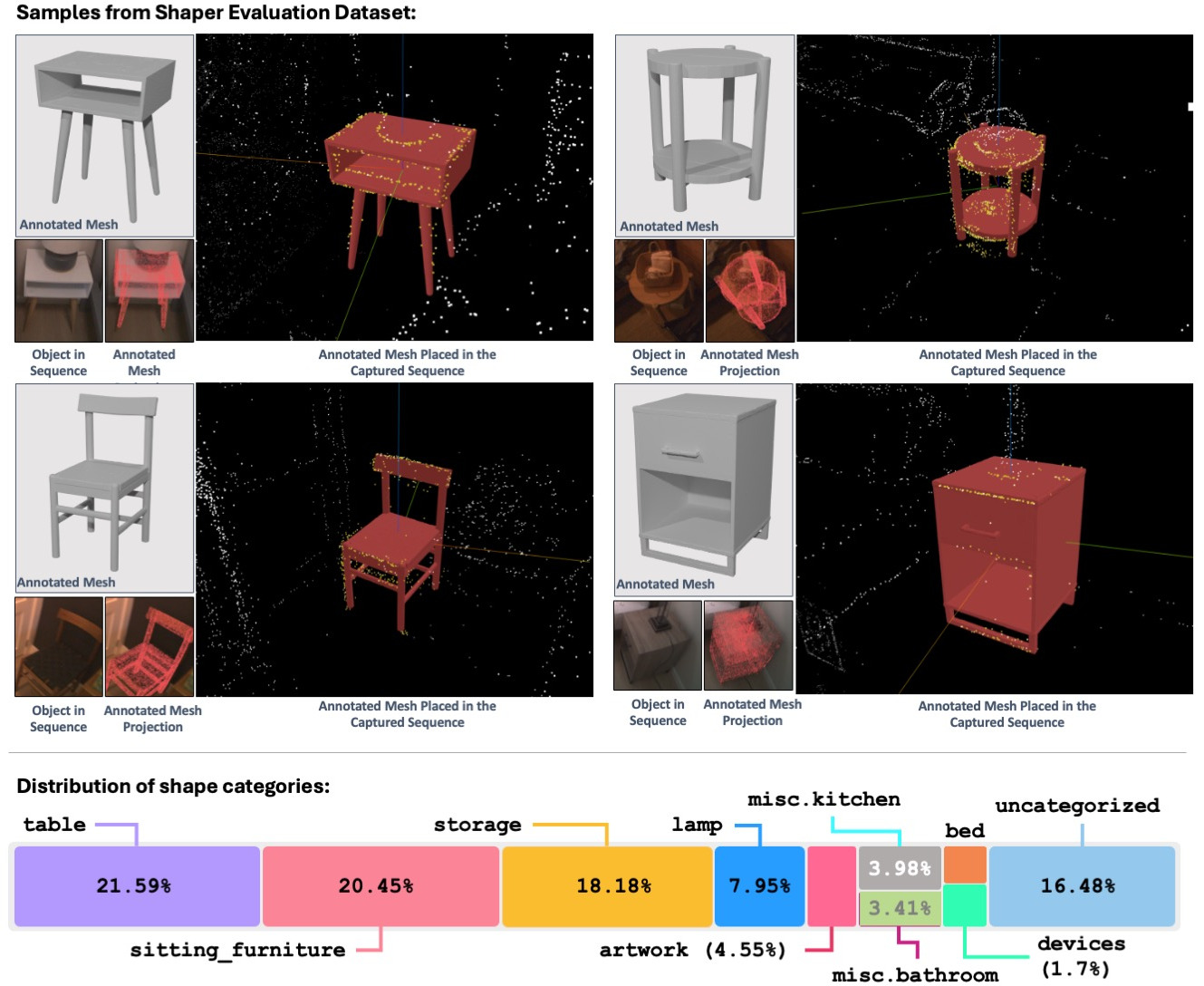}
   \vspace{-4mm}
   \caption{(Top) Examples from the ShapeR evaluation dataset. Each sub-image shows the annotated ground-truth mesh, a representative frame containing the object, the mesh placed within the sequence, and the projection of the mesh onto the image. (Bottom) Distribution of object shapes categories in the ShapeR evaluation set, covering 178 objects across 7 sequences}
   \label{fig:dataset_example}
\end{figure}

%% file: figures/dataset_annotation.tex
\begin{figure*}
  \centering
   \vspace{-7mm}
   \includegraphics[width=\linewidth]{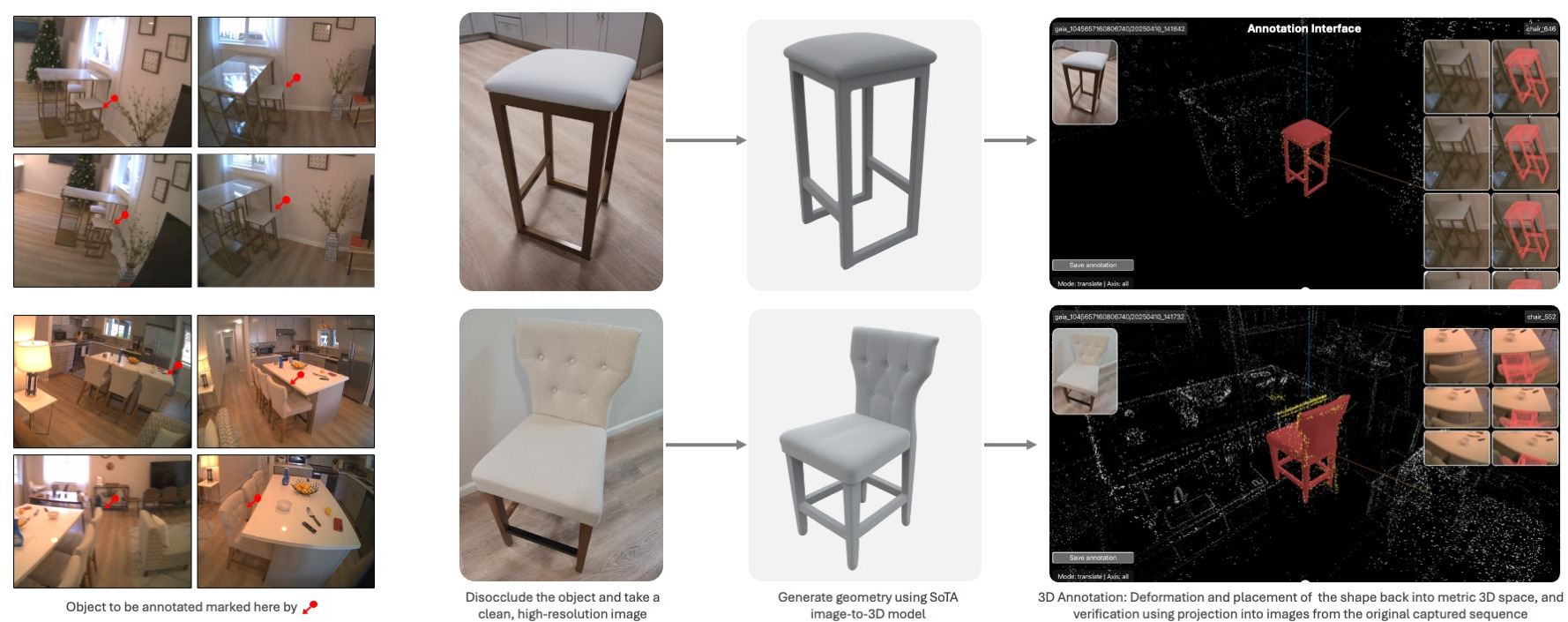}
   \caption{To obtain pseudo-ground truth geometry for an object in the sequence (left), we first place the object in isolation to avoid clutter and occlusion, and capture a high-quality, uncluttered image. We then apply segmentation and image-to-3D modeling to generate the object's geometry (mid). This geometry is manually aligned and inserted back into the original casual sequence using a web annotation interface, verified by matching 2D projections to image silhouettes and by checking alignment with the sequence’s point cloud (right).}
   \label{fig:dataset_annotation}
   \vspace{-2mm}
\end{figure*}

%% file: figures/comparison_lirm_graph.tex
\begin{figure}
  \centering
   \includegraphics[width=\linewidth]{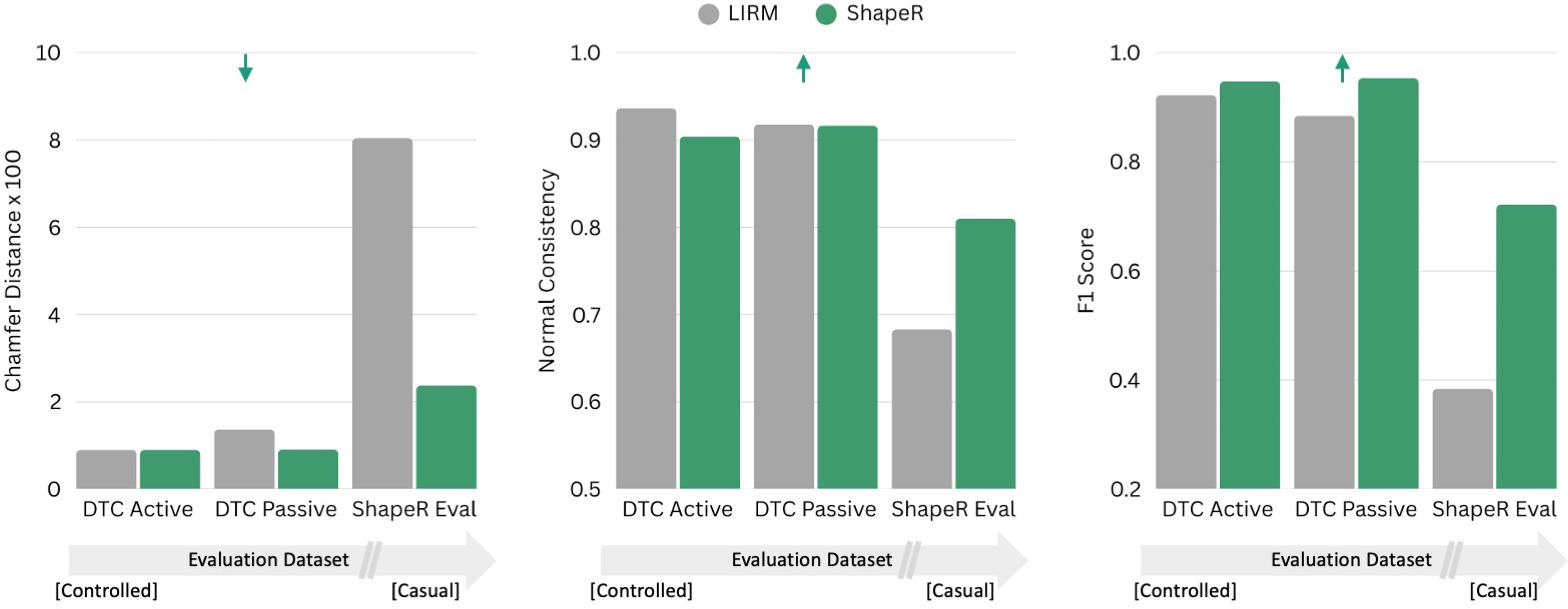}
   \caption{DTC Active, DTC Passive, and ShapeR Evaluation datasets represent a progression from highly controlled capture setups (DTC Active), to slightly less controlled environments (DTC Passive), and finally to casual, real-world scenes (ShapeR Evaluation). As the datasets become more challenging, baseline method metrics deteriorate, while ShapeR remains comparatively stable. Notably, the increase in scene casualness is not linear; ShapeR Evaluation is significantly more challenging than DTC Passive.}
   \label{fig:comparison_trend}
\end{figure}

%% file: figures/mapanything.tex
\begin{figure}[tp]
  \centering
   \includegraphics[width=\linewidth]{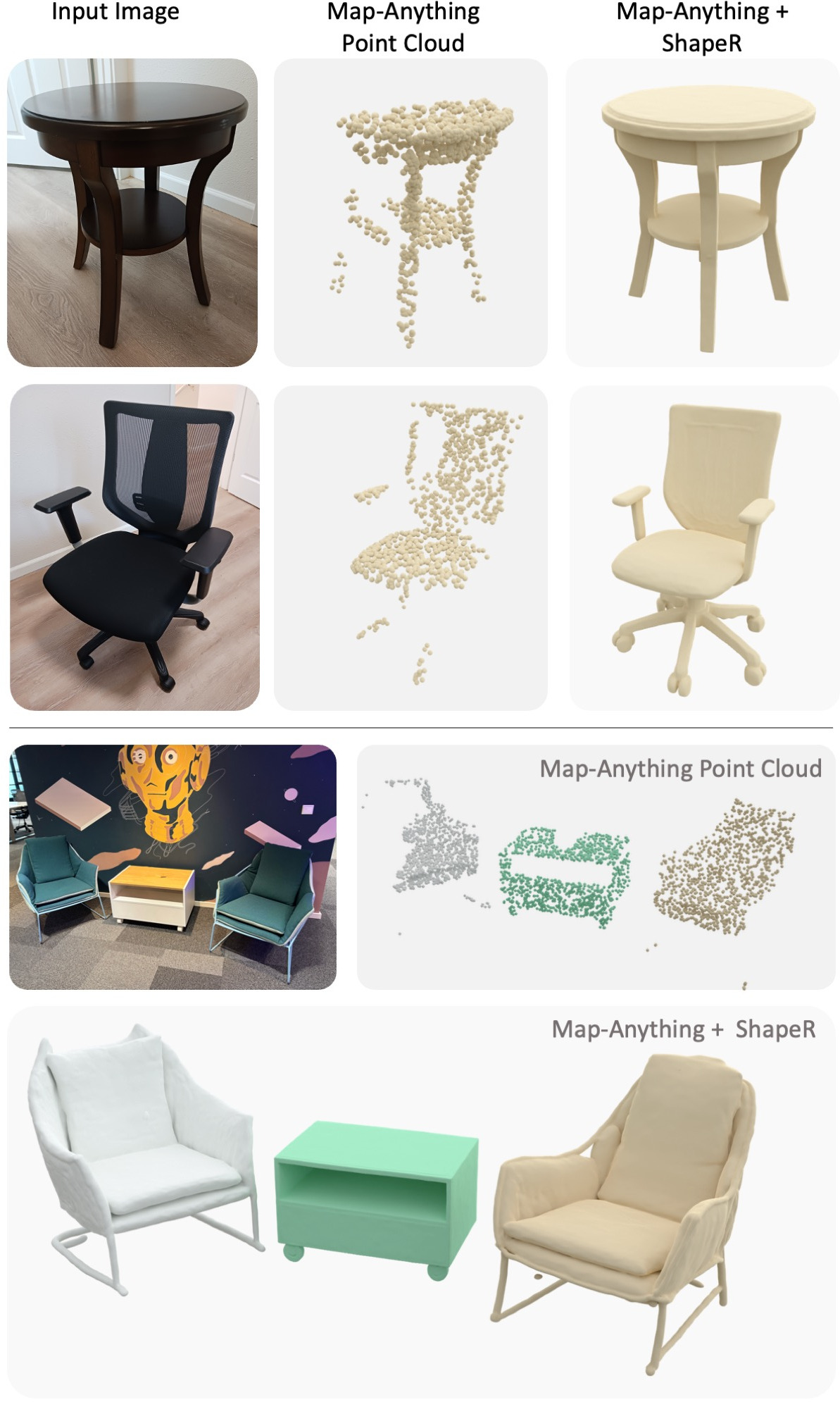}
   \caption{Single image to metric 3D with ShapeR. While ShapeR is trained to leverage posed multi-view signals, it can be configured for single-image 3D reconstruction without retraining by using a metric point cloud and camera estimator such as MapAnything \cite{keetha2025mapanything}. This enables ShapeR to generate metrically accurate 3D shapes from a monocular image.}
   \label{fig:map_anything}
\end{figure}

%% file: figures/limitations.tex
\begin{figure}[tp]
  \centering
   \includegraphics[width=\linewidth]{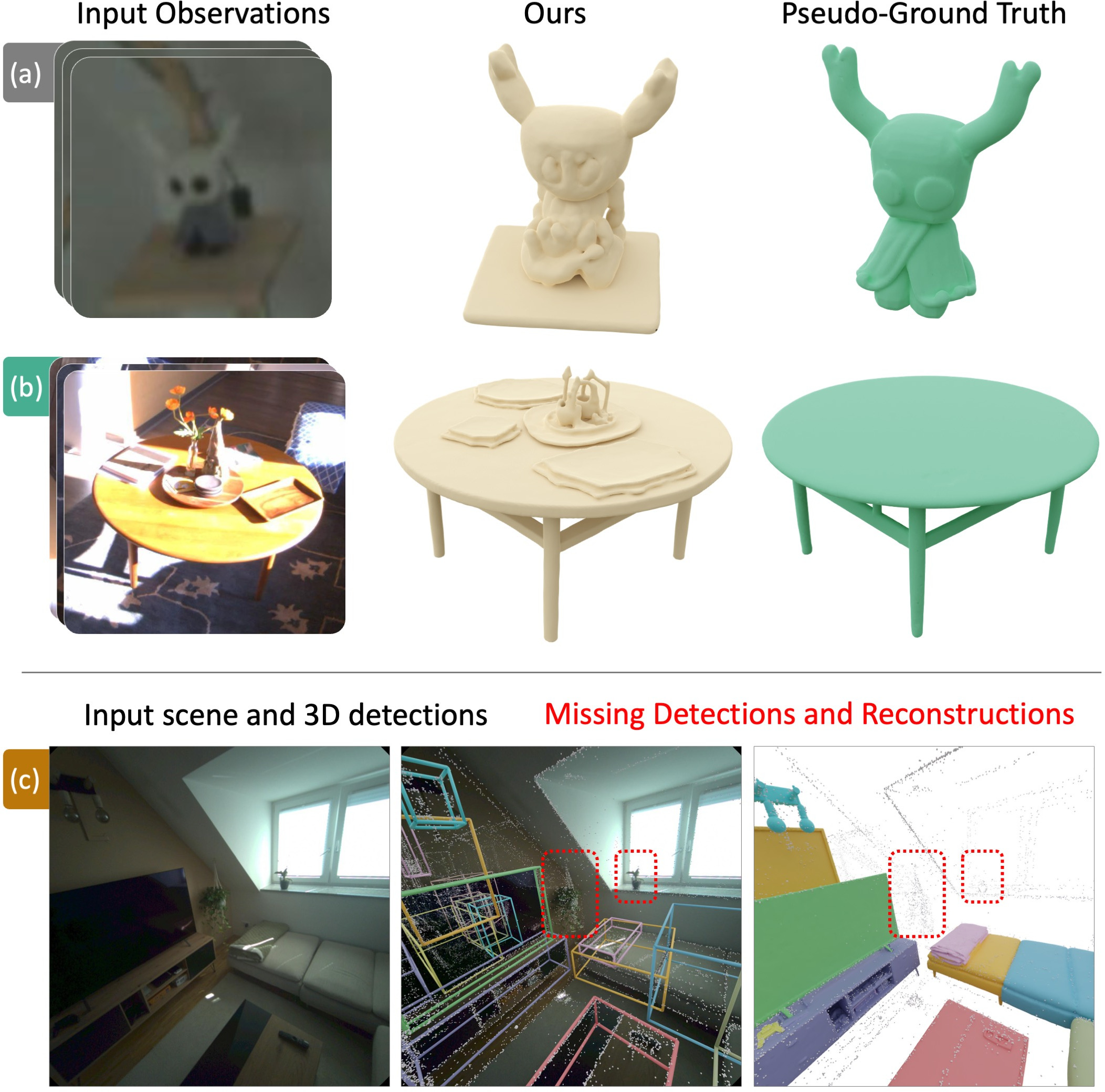}
   \caption{ShapeR limitations. (a) Low image fidelity or limited views lead to incomplete or low-detail reconstructions. (b) Closely stacked or attached objects can cause meshes to include parts of adjacent structures, even when the point associated with these structures are not in the input (c) ShapeR relies on upstream 3D detection; missed or inaccurate detections result in unrecoverable objects.}
   \vspace{-6mm}
   \label{fig:limitations}
\end{figure}

%% file: figures/comparison_sam3d.tex
\begin{figure*}
  \centering
   \includegraphics[width=0.95\linewidth]{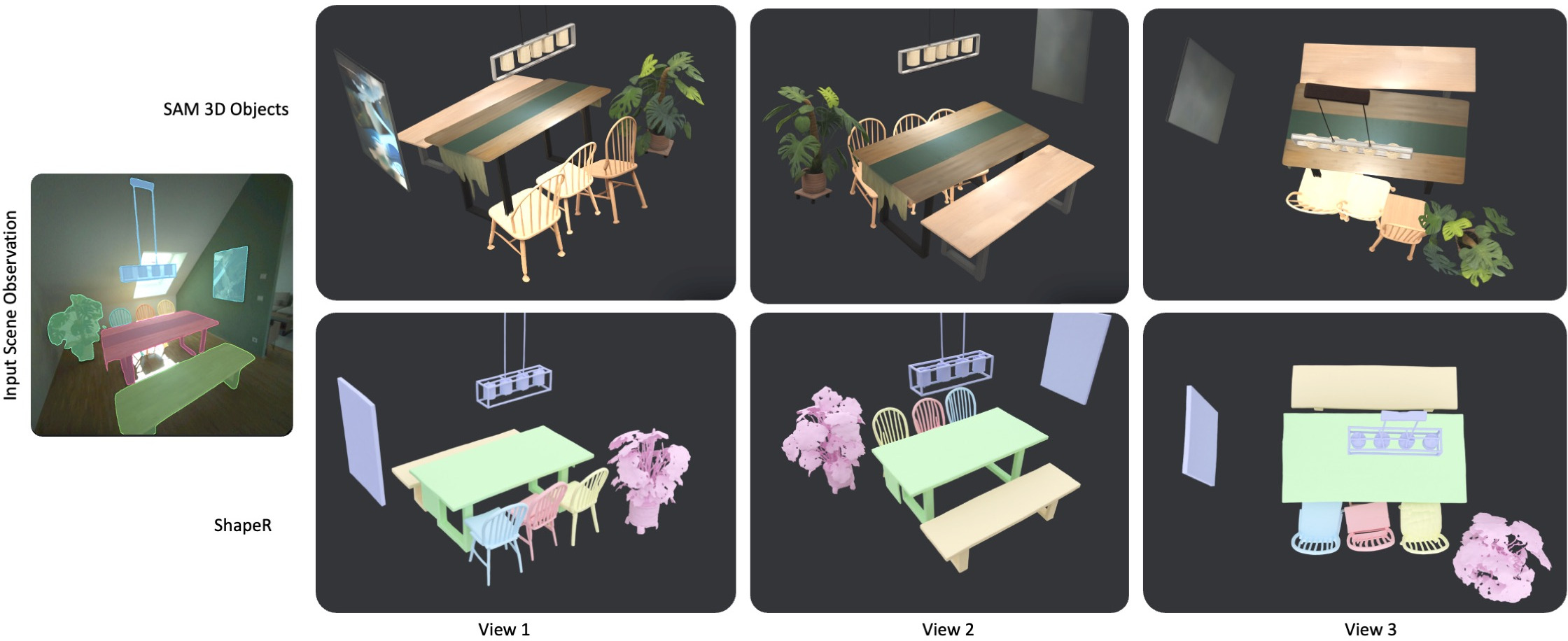}
   \caption{Comparison with SAM 3D Objects \cite{sam3dobjects}. SAM 3D Objects takes a single image and interactive object segments to produce non-metric 3D shapes, which are generally accurate but may exhibit minor hallucinations (e.g., predicting five lamps instead of four) and poor object placement. In contrast, ShapeR leverages posed images from a sequence to generate metrically accurate geometry and consistently well-placed objects.}
   \label{fig:comparison_sam3d}
\end{figure*}

%% file: figures/comparison_lirm_qualitative.tex
\begin{figure*}
  \centering
   \includegraphics[width=\linewidth]{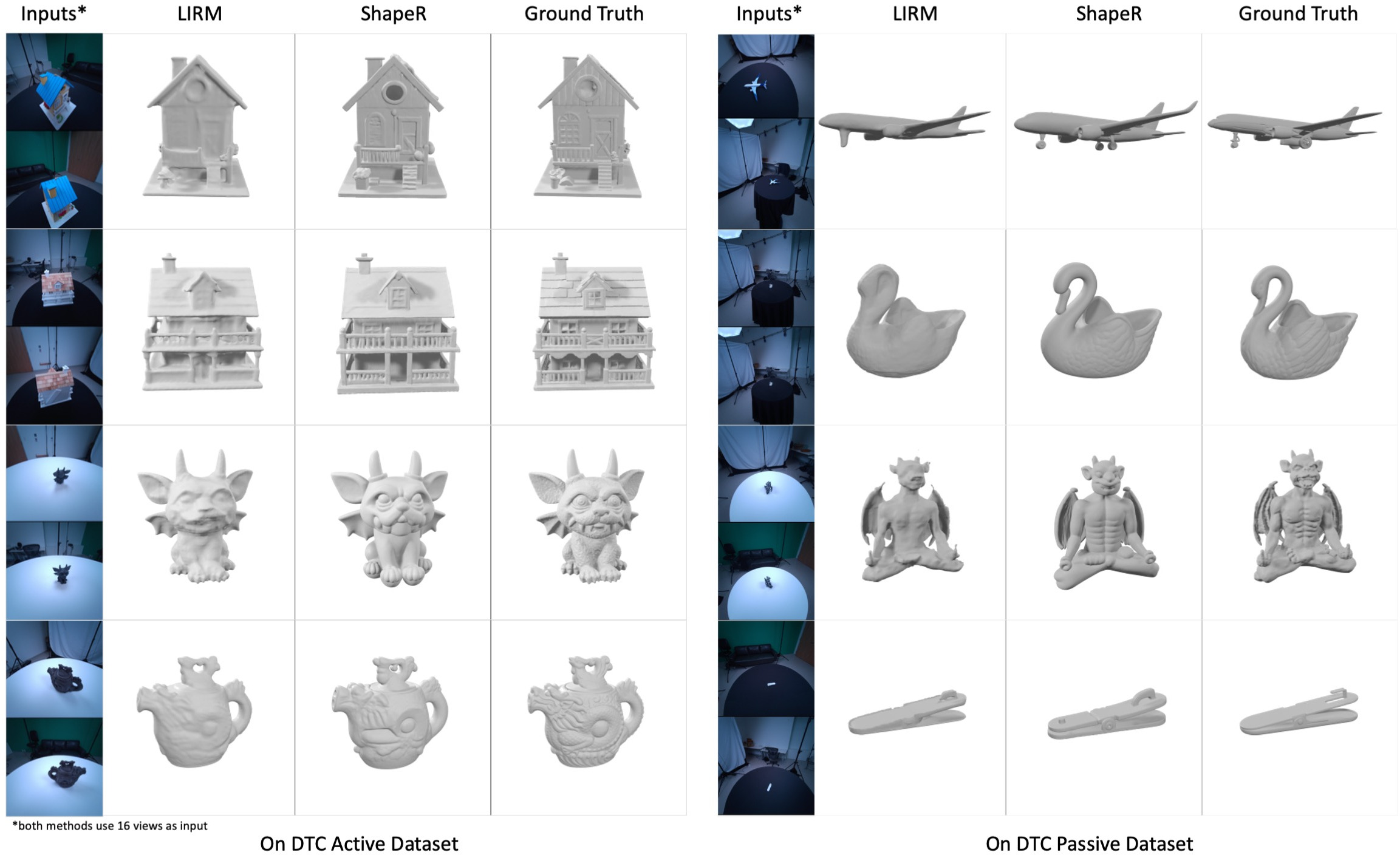}
   \caption{Comparison against LIRM \cite{li2025lirm} on DTC \cite{dong2025digital} Active and Passive sequences. Both setups feature tabletop objects without clutter or occlusions; however, Passive sequences allow more free user movement, while Active sequences involve the user circling the object. ShapeR performs competitively on Active sequences and surpasses LIRM on the slightly more casual Passive sequences.}
   \label{fig:comparison_dtc_lirm}
\end{figure*}

%% file: figures/result_thirdparty.tex
\begin{figure*}
  \centering
   \includegraphics[width=\linewidth]{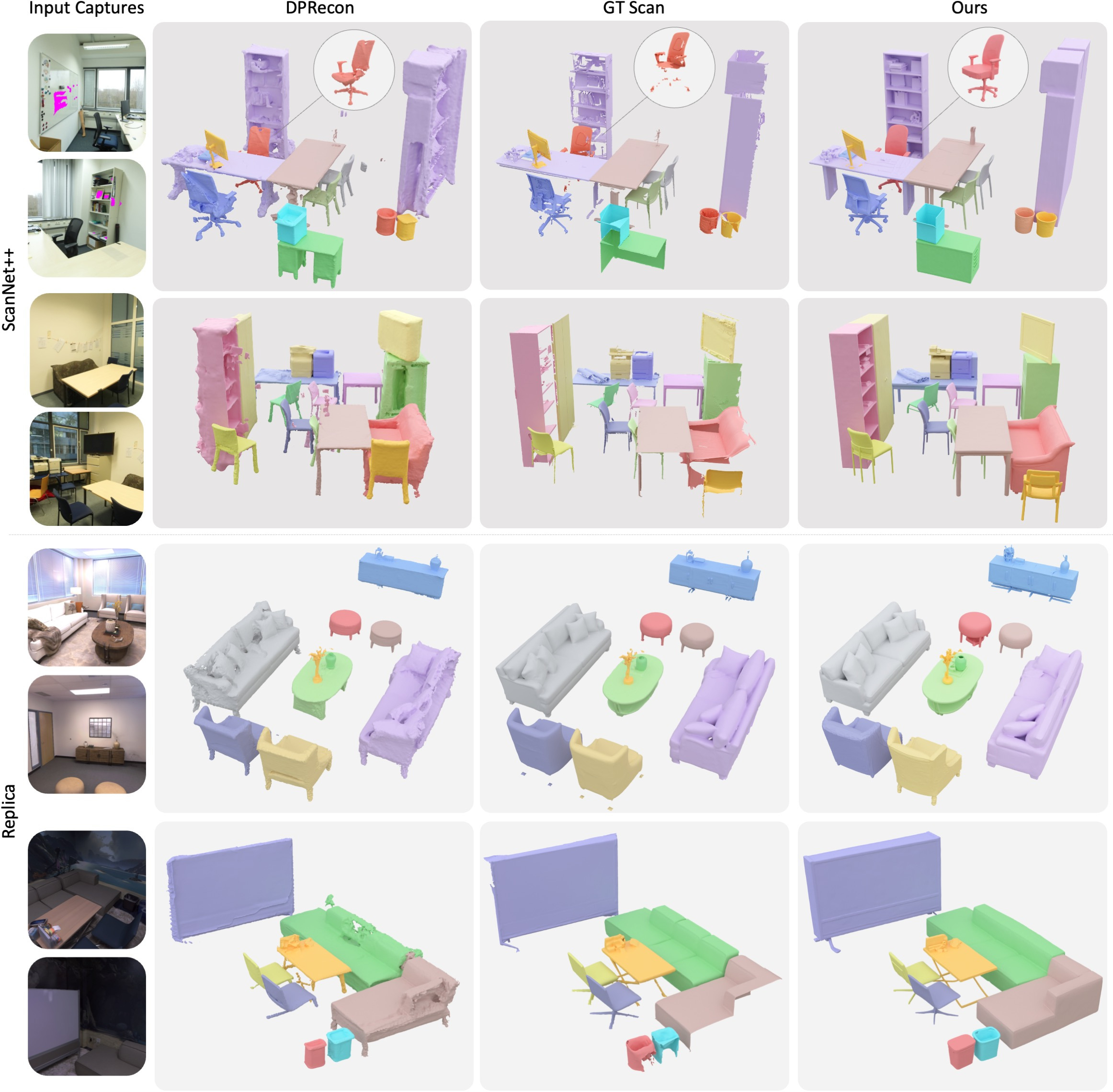}
   \caption{Reconstruction results on ScanNet++ \cite{yeshwanth2023scannet++} and Replica \cite{straub2019replica} scenes, compared to DPRecon \cite{ni2025decompositional}. ShapeR produces complete reconstructions, often surpassing the ground-truth scans in completeness, as the latter lack geometry in occluded regions.}
   \label{fig:results_thirdparty}
\end{figure*}

%% file: tables/thirdparty.tex
\begin{table}
\caption{Reconstruction performance comparison on ScanNet++~\cite{yeshwanth2023scannet++} and Replica~\cite{straub2019replica} datasets against DPRecon~\cite{ni2025decompositional}. We use six scenes from ScanNet++ and seven scenes from Replica as processed by DPRecon. Note that chamfer distance, normal consistency and recall (R) are calculated in one direction, \ie only point present on ground truth meshes are used for evaluation, due to tha lack of incomplete meshes present in these datasets.}
\label{tab:comparison_thirdparty}
\centering
\resizebox{\linewidth}{!}
{
\begin{tabular}{@{}l ccc ccc @{}}
\toprule
\multirow{2}{*}{Methods} & \multicolumn{3}{c}{ScanNet++} & \multicolumn{3}{c}{Replica} \\
\cmidrule(lr){2-4} \cmidrule(lr){5-7} 
 & CD$\times10^2\downarrow$ & NC$\uparrow$ & R$\uparrow$ & CD$\times10^2\downarrow$ & NC$\uparrow$ & R$\uparrow$ \\
\midrule
DPRecon~\cite{ni2025decompositional} & 7.69  & 0.73 & 0.45 & 4.65 & 0.75  & 0.57  \\
\OURS & \textbf{1.09} & \textbf{0.84} & \textbf{0.91} & \textbf{1.77} & \textbf{0.84} & \textbf{0.82} \\
\bottomrule
\end{tabular}
}
\end{table}

%% file: tables/comparison_dtc.tex
\begin{table}
\caption{Reconstruction results on the DTC \cite{dong2025digital} Active and Passive datasets, each with approximately 100 sequences, compared against LIRM \cite{li2025lirm}. ShapeR achieves comparable performance to LIRM on the highly controlled Active sequences, and surpasses LIRM on the more challenging Passive sequences.}
\label{tab:comparison_dtc}
\centering
\resizebox{\linewidth}{!}
{
\begin{tabular}{@{}l ccc ccc @{}}
\toprule
\multirow{2}{*}{Methods} & \multicolumn{3}{c}{DTC Active} & \multicolumn{3}{c}{DTC Passive} \\
\cmidrule(lr){2-4} \cmidrule(lr){5-7} 
 & CD$\times10^2\downarrow$ & NC$\uparrow$ & F1$\uparrow$ & CD$\times10^2\downarrow$ & NC$\uparrow$ & F1$\uparrow$ \\
\midrule
LIRM~\cite{li2025lirm} & \textbf{0.90} & \textbf{0.94} & 0.92 & 1.37 & 0.91 & 0.88  \\
\OURS & 0.94 & 0.91 & \textbf{0.94} & \textbf{0.95} & 0.91 & \textbf{0.95} \\
\bottomrule
\end{tabular}
}
\end{table}